\documentclass{article}

\usepackage{PRIMEarxiv}

\usepackage[utf8]{inputenc} 
\usepackage[T1]{fontenc}    
\usepackage{hyperref}       
\usepackage{url}            
\usepackage{booktabs}       
\usepackage{amsfonts}       
\usepackage{nicefrac}       
\usepackage{microtype}      
\usepackage{lipsum}
\usepackage{graphicx}
\graphicspath{{media/}}     
\usepackage{amsmath,amsfonts,amssymb}
\usepackage{rotating}
\usepackage{graphicx}
\usepackage{setspace}
\usepackage{tocloft}
\usepackage{multirow}
\usepackage{makecell}
\usepackage{listings}
  
\title{AI-Generated Fall Data: Assessing LLMs and Diffusion Model for Wearable Fall Detection}

\author{
  Sana Alamgeer \\
  Texas State University \\
  San Marcos, USA \\
  \texttt{sanaalamgeer@gmail.com} \\
  \And
  Yasine Souissi \\
  University of North Carolina \\
  Charlotte, USA \\
  \texttt{yasinesouissi@gmail.com} \\
   \And
  Anne H. H. Ngu \\
  Texas State University \\
  San Marcos, USA \\
  \texttt{angu@txstate.edu} \\
}

\begin{document}
\maketitle

\begin{abstract}
Training fall detection systems is challenging due to the scarcity of real-world fall data, particularly from elderly individuals. To address this, we explore the potential of Large Language Models (LLMs) for generating synthetic fall data. This study evaluates text-to-motion (T2M, SATO, ParCo) and text-to-text models (GPT4o, GPT4, Gemini) in simulating realistic fall scenarios. We generate synthetic datasets and integrate them with four real-world baseline datasets to assess their impact on fall detection performance using a Long Short-Term Memory (LSTM) model. Additionally, we compare LLM-generated synthetic data with a diffusion-based method to evaluate their alignment with real accelerometer distributions. Results indicate that dataset characteristics significantly influence the effectiveness of synthetic data, with LLM-generated data performing best in low-frequency settings (e.g., 20Hz) while showing instability in high-frequency datasets (e.g., 200Hz). While text-to-motion models produce more realistic biomechanical data than text-to-text models, their impact on fall detection varies. Diffusion-based synthetic data demonstrates the closest alignment to real data but does not consistently enhance model performance. An ablation study further confirms that the effectiveness of synthetic data depends on sensor placement and fall representation. These findings provide insights into optimizing synthetic data generation for fall detection models.
\end{abstract}


\section{Introduction}\label{sect:intro}
According to a report published by the World Population Prospects~\cite{un_world_2022}, the number of people aged 65 and older is projected to double by 2050, which will surpass the number of children under the age of 5. Another report~\cite{who_ageing_health} published by the World Health Organization (WHO) states that the global proportion of individuals aged 60 and above is expected to increase from 11\% to 22\%, growing from around 605 million to nearly 2 billion over the coming decades. This demographic shift brings major health concerns for the elderly, with falls being a significant issue. Therefore, it is crucial to develop effective measures to monitor and manage falls.

\begin{figure}[tbh!]
\vspace{-10px}
  \centering
  \includegraphics[width=0.7\textwidth]{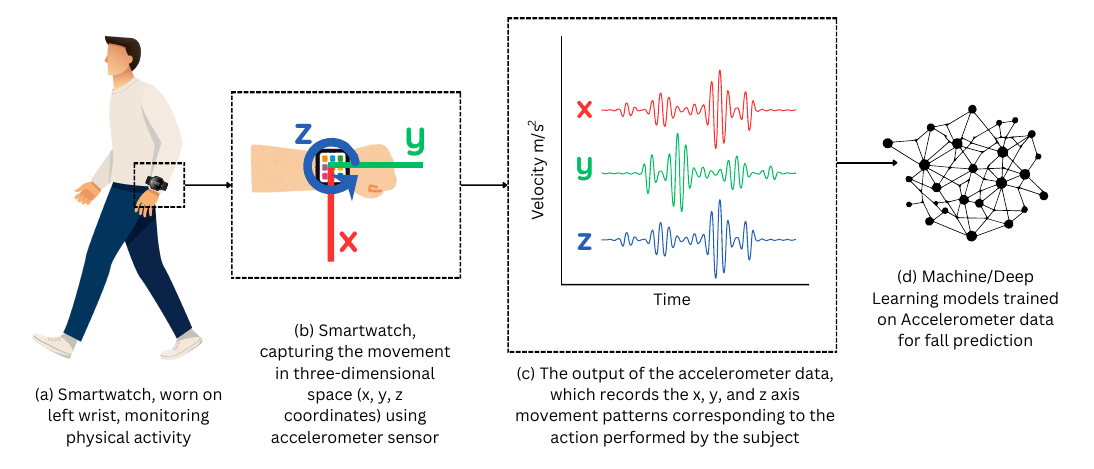}
  \caption{Illustration of smartwatch-based fall detection: (a) A person wearing a smartwatch equipped with inertial sensors, which continuously monitor movement. (b) Representation of the internal accelerometer of the smartwatch, capturing movement along three axes: $x$ (red), $y$ (green), and $z$ (blue). (c) The accelerometer data, corresponding to the $x$, $y$, and $z$-axes, are visualized as time-series signals. (d) This data is then processed by machine/deep learning algorithms to detect patterns indicating a fall.}
  \label{fig:watch_and_accelerometer}
\end{figure}

In response to this need, fall detection systems based on wearable sensors, known as wearable sensors-based human activity recognition (WSHAR), have become essential tools in elderly care to prevent serious injuries and enable timely medical intervention. Wearable sensors contain inertial measurement units (IMU), primarily accelerometers, that generate accelerometer data capturing the wearer’s movements in real-time. This accelerometer data is used by machine/deep learning models for training, and deployed on wearable devices, such as smartphones or smartwatches, to continuously monitor the user's activity and detect falls as they occur. For this purpose specifically, smartwatches have emerged as the most comfortable and practical solution for continuous monitoring, as they can unobtrusively collect data on the wearer’s movements~\cite{demo_persona_fall}. Elderly people, in particular, are more likely to wear these devices consistently, which makes them ideal for real-time fall detection. Figure~\ref{fig:watch_and_accelerometer} illustrates the application of smartwatch-based fall detection. 

To ensure accurate fall detection and minimal false alarms, a substantial amount of accelerometer data, representing various fall scenarios, must be collected for training the models. However, collecting real fall data from elderly individuals is challenging, because it is both impractical and ethically problematic due to the risks involved. Consequently, researchers often make use of fall data generated by younger individuals to train these models. This approach, however, introduces significant discrepancies in model performance, as the movement patterns and accelerometer data produced by young and elderly people may differ.
The mismatch in data characteristics often leads to a high rate of false positives, which results in reducing the reliability and effectiveness of fall detection systems when deployed in real-world elderly settings.

To address this challenge, generating synthetic data presents a potential solution to simulate fall scenarios that closely resemble the movements of elderly individuals. One promising approach involves using Large Language Models (LLMs) to generate synthetic data. These models have demonstrated success in other healthcare domains, such as patient monitoring and activity recognition~\cite{liu2023fewshot, leng2023, zhou2024, xu2024, wang2024}, by leveraging their advanced reasoning and data generation capabilities. Given the semantics of pre-trained LLMs, it is worth exploring their potential to generate synthetic fall data. To the best of our knowledge, no prior research has systematically investigated the use of LLMs for synthetic fall data generation and its subsequent impact on fall detection systems.

It is important to acknowledge the absence of public domain fall data specifically collected from elderly individuals, which could otherwise be leveraged to fine-tune models for a more accurate representation of real-world fall dynamics in this demographic. Generating age-inclusive fall data is one of our future endeavors. Our focus in this paper is therefore not on generating age-specific movement patterns but on exploring the capability of LLMs to synthesize accelerometer data for generic fall simulations, assessing its impact on a fall detection model trained with a combination of real and synthetic falls, and most importantly, setting the foundation for future advancements in generating elderly-specific fall data. Building on this motivation, this study aims to address three critical research questions: 
\begin{itemize}
    \item Can LLMs generate accelerometer data specific to gender, age, or joint?
    \item Do LLM-generated data perform better or align more closely with real data than diffusion-based methods?
    \item Do LLM-generated data improve the performance of the model for fall detection tasks?    
\end{itemize}

By systematically exploring these questions, we aim to provide insights into the potential and limitations of using LLMs for generating synthetic data in the context of fall detection, which will ultimately guide future research and practical implementations in this area. Our major contributions are as follows:
\begin{itemize}
    \item Data Generation: First, we generate synthetic fall data using two categories of pre-trained LLMs: text-to-motion and text-to-text generation models. These models receive prompts describing fall scenarios and produce accelerometer data accordingly.
    
    \item Qualitative and Quantitative Analysis: We conduct both qualitative and quantitative analyses to evaluate the distribution of synthetic data in comparison to real data, and assess the degree of alignment between the generated and real data.
    
    \item We investigate the impact of augmenting real fall data with synthetic fall data in training a Long Short-Term Memory (LSTM)-based fall detection model. The performance of the augmented model is then compared against a baseline model trained exclusively on real data.
    
    \item Ablation Study: Finally, we conduct an ablation study to evaluate the impacts of the quantity of synthetic data, prompting techniques, and baseline dataset characteristics when combined with synthetic data.
\end{itemize}

The remainder of this document is organized as follows: In Section~\ref{sect:related_work}, we discuss the related work, providing an overview of the existing literature on synthetic data generation for the systems of fall detection. Section~\ref{sect:method} details the methodology used in this study, including the synthetic data generation techniques and the machine learning models employed. Section~\ref{sect:implement_detaisl} covers the implementation details and experimental setup, including data preprocessing and model evaluation. Section~\ref{sect:results_disc} presents the results and discussion, addressing three main research questions. Section~\ref{sect:ablation_study} provides an ablation study to validate the key findings, and Section~\ref{sect:key_findings} summarizes the key findings of the study, where we summarize the insights on the effectiveness of LLM-generated synthetic data in enhancing fall detection model. The document concludes with Section~\ref{concusion_fw}, where we present a summary of this work and propose directions for future work.

\section{Related Work}\label{sect:related_work}
In this section, we review synthetic data generation techniques for human activity recognition (HAR) and fall detection, covering both non-generative AI and generative AI techniques, and highlighting their successes and limitations. 

\subsection{Non-generative AI Techniques}
\textbf{Deep Learning-based Techniques:} Alharbi et al.~\cite{9206624} investigate the use of Wasserstein Generative Adversarial Networks (WGANs) to generate synthetic sensor data to address the class imbalance in HAR datasets. The authors highlight the challenge of collecting large, balanced datasets for deep learning-based HAR classifiers and propose using WGANs to generate synthetic data for under-represented activity classes. Their findings indicate that oversampling imbalanced datasets with WGAN-generated synthetic data can significantly enhance the performance of 1-dimensional convolution neural network (1D-CNN) classifiers, improving F1-scores by 7\% to 10\%.

Liaquat et al.~\cite{Majid2023} create synthetic sensor data that mimics the baseline data collected from wearable sensors for 12 different activities. Three deep learning-based techniques are used for synthetic data generation: Synthetic Data Vault Probabilistic Autoregressive (SDV-PAR)~\cite{par2022sequential}, Time-series Generative Adversarial Network (TGAN)~\cite{yoon2019time}, and Conditional Tabular Generative Adversarial Network (CTGAN)~\cite{xu2019modeling}. The findings show that while real data achieved an accuracy score of 0.9725, the synthetic data generated by CTGAN achieved a lower accuracy score of 0.8373. This indicates that synthetic data can be a viable solution, though it may not yet match the performance of baseline data.

Matthews et al.~\cite{matthews2020} create a large synthetic dataset of videos depicting actions like walking, waving, and sitting down relying on 3D rendering tools. They train an I3D model~\cite{I3D2018} on this synthetic dataset and evaluate its performance on the baseline dataset. The key finding is that combining their synthetic dataset with the baseline training dataset leads to a 2\% improvement in the performance of the classification model. 

Another method that relies on 3D rendering tools, is ElderSim proposed by Hwang et al.~\cite{hwang2020}, which generates synthetic data of elders, performing activities of daily living, to train deep learning models for HAR in eldercare applications. The authors simulate 3D realistic motions of characters in various environments with adjustable data-generating options, including viewpoints and lighting. The generated data alongside baseline datasets show that augmenting training data with synthetic data leads to a noticeable improvement in action recognition performance. 

The methods described above have shown promising results in generating synthetic data for HAR-related tasks. However, these approaches have limitations. For example, WGANs and other GAN-based techniques, while effective in addressing class imbalance, often require extensive tuning and can be sensitive to the quality of the initial data, potentially leading to poor generalization. Additionally, methods relying on 3D rendering tools, like ElderSim, are computationally expensive and may struggle to capture the full complexity of real-world movements, particularly in diverse and uncontrolled environments.

\textbf{Pose Estimation Techniques:} An alternative approach for generating time-series data involves extracting motion information from video recordings using pose estimation techniques\cite{Singh2022, Dubey2023, Stenum2024}. By analyzing sequential video frames, these methods track human movements over time, converting them into time-series representations that capture joint angles, velocities, and other kinematic variables. However, the effectiveness of this approach heavily depends on the quality of video data and the accuracy of the pose estimation algorithms. Factors such as occlusions, variations in lighting conditions, and individual differences in movement patterns can introduce errors and inconsistencies, ultimately impacting the reliability of the extracted time-series data\cite{Tang2023}.

\subsection{Generative AI}
\textbf{Text-to-Motion Methods:} In the context of fall detection and human activity recognition, there has been limited research in the state-of-the-art focusing on the use of Generative AI for synthetic data generation. One of these works is proposed by Leng et al.~\cite{leng2023har}, which introduces a novel approach to tackle the issue of limited annotated data in HAR using GPT-4~\cite{openai2023gpt4} and text-driven motion synthesis model T2M-GPT~\cite{t2mgpt2023}. The method involves generating virtual IMU data from text descriptions. The authors demonstrate that synthetically generated data significantly enhances downstream classifier performance across other benchmark datasets. This work primarily focuses on text-driven motion synthesis and does not fully capture the complexities and variabilities of real-world fall scenarios, particularly those involving elderly individuals.

Meng et al.~\cite{meng2024} proposed a text-driven human motion generation model that integrates insights from Vector Quantization (VQ)-based approaches~\cite{t2mgpt2023, vqvae2018, guo2023momask} to optimize diffusion-based method~\cite{NEURIPS2020_4c5bcfec} for human motion generation. It performs bidirectional masked autoregression, improving data representation and distribution1. However, it still faces challenges in maintaining diversity and avoiding information loss. Chi et al.~\cite{chi2024m2d2m} proposed a multi-motion generation model from text with a diffusion model~\cite{NEURIPS2020_4c5bcfec}. It generates human motion sequences from textual descriptions of multiple actions, ensuring seamless transitions and coherence through a dynamic transition probability and a two-phase sampling strategy (first jointly sampling a coarse outline of multi-motion sequences, then independently sampling to refine each motion). Its limitation lies in the complexity of accurately modeling diverse motion sequences. Wu et al.~\cite{wu2024mote} proposed a text-to-motion diffusion model for multiple generation tasks. It is a unified multi-modal model that handles paired text-motion generation, motion captioning, and text-driven motion generation by learning the marginal, conditional, and joint distributions of motion and text. Its main limitation is the dependency on high-quality data for training, which can affect performance in diverse real-world scenarios.

Another method, although not targeting HAR synthetic data, is presented by Tang et al.~\cite{tang2023synthetic}, which investigates the use of LLMs to generate synthetic data for enhancing clinical text mining. The authors explore how this synthetic data can improve tasks like biological named entity recognition (NER) and relation extraction (RE). Their findings indicate significant performance improvements, with the F1-score for NER, increasing from 23.37\% to 63.99\% and for RE from 75.86\% to 83.59\%. The improvements observed in tasks like biological NER and RE do not necessarily translate to the challenges of generating synthetic sensor data for HAR, where the physical and temporal dynamics of human motion are crucial. 

\textbf{Digital Twins:} This is an emerging approach for generating synthetic data through precise simulations of physical systems. Recent studies~\cite{tang2023synthetic, Tufisi2024} have utilized this method to simulate forward fall dynamics, creating synthetic inertial data for training machine learning models. This technique enables the controlled generation of diverse datasets, mitigating the ethical and logistical challenges associated with real-world data collection. However, digital twin-based simulations require extensive calibration to accurately reflect real-world variability and face challenges in generalizing across diverse populations and sensor configurations, limiting their applicability in heterogeneous settings.

\section{Methodology}\label{sect:method}
In this section, we provide a comprehensive overview of the methodologies employed in this study. First, we describe the synthetic data generation techniques, encompassing both text-to-motion and text-to-text generation methods in section~\ref{subsect:data_gen}. The subsequent section~\ref{subsubsect:data_prep} outlines the preprocessing steps applied to the datasets, ensuring they are appropriately structured for model training and testing. Finally, in section~\ref{subsect:lstm}, we briefly explain the architecture of the falls classification model utilized in this study.

\subsection{Synthetic Data Generation}\label{subsect:data_gen}
\begin{figure}[tbh!]
  \centering
  \includegraphics[width=1.0\textwidth]{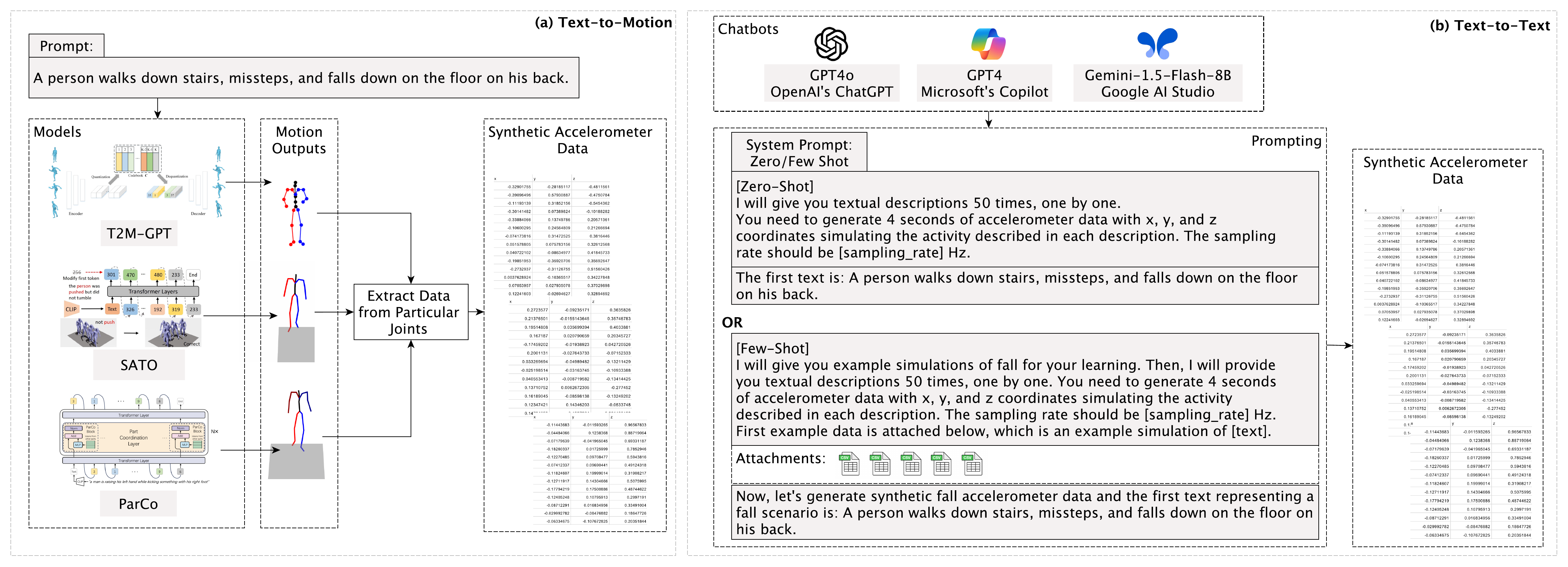}
  \caption{Overview of the data generation process using two categories of pre-trained large language models (LLMs). (a) Text-to-Motion: Motion data is generated using models, T2M-GPT, SATO, and ParCo, and then, joint-specific data is extracted to create synthetic accelerometer data. (b) Text-to-Text: Three LLMs, GPT4o (ChatGPT), GPT4 (Copilot), and Gemini-1.5-Flash-8B (Google AI Studio), take prompts to simulate fall scenarios and generate accelerometer data directly for the left wrist.}
  \label{fig:data_generation}
\end{figure}
This section discusses the generation of synthetic accelerometer data using two categories of pre-trained large language models (LLMs). As shown in Figure~\ref{fig:data_generation}(a) and (b), the first category is ``text-to-motion'', which involves using prompts representing fall scenarios to generate motion data. Then, from this motion data, we extract joint-specific data (including left/right wrist, waist/pelvic). The second category, ``text-to-text generation'', consists of using system prompts to set directions, followed by generating accelerometer data with $x$, $y$, and $z$ coordinates using the same prompts that reflect fall scenarios. The detailed process for each approach is provided below.

\subsubsection{Text-to-Motion Generation}\label{subsubsect:text2motion}

This category involves generating synthetic motion data using pre-trained LLMs that are specifically designed for motion synthesis. In this approach, we initially designed a list of 50 prompts based on insights gained from 200 iterative attempts, ultimately selecting the prompts that produced the most accurate fall motion sequences. Then, we feed these prompts one by one into three motion generation models selected based on their relevance to human motion generation, recent advancements in the field, and ease of use. These models are: Generating Human Motion from Textual Descriptions with Discrete Representations (T2M-GPT)\cite{t2mgpt2023}, Stable Text-to-Motion Framework (SATO)\cite{sato2024}, and Part-Coordinating Text-to-Motion Synthesis (ParCo)~\cite{parco2024}, which take a descriptive text as input and produce motion data involving multiple joints of the human body.

The T2M-GPT (T2M) model leverages the power of Generative Pre-trained Transformer (GPT)~\cite{gpt2018} architecture to produce human motion based on textual descriptions. The model utilizes a Vector Quantized-Variational Autoencoder (VQ-VAE)\cite{vqvae2018} to quantize the motion data into discrete codes, which are then decoded into motion sequences. The ParCo model introduces an innovative approach to text-to-motion generation by focusing on part-based motion generation. Unlike T2M-GPT and SATO, which generate whole-body motions, ParCo decomposes the body into six parts (e.g., right arm, left leg, backbone) and uses multiple lightweight generators to create motions for each part. These part motions are then coordinated through a Part Coordination module, ensuring that the generated motions are not only fine-grained but also coordinated and realistic. Both T2M and ParCo use transformer-based approaches that focus on the discrete representation of motion. SATO utilizes a diffusion model~\cite{motiondiffus2022}, which is particularly effective in generating high-quality, temporally coherent motion sequences, thereby reducing jitter and instability often observed in other generative models. T2M, ParCo and SATO are trained on the HumanML3D\cite{humanml3D2022} and KIT-ML~\cite{kitmotion2016} datasets, which contain a substantial number of 3D human motion sequences paired with textual descriptions.

These models generate motion data in \textit{.npy} format, from which, we extract joint-specific data, located at different indices (Figure~\ref{fig:data_generation}(a)), to ensure it aligns with our baseline datasets. Once the relevant data is retrieved, it is stored in a CSV file, formatted with semicolon-separated values, and organized under the headers \textit{x}, \textit{y}, and \textit{z} to represent the respective coordinates.

\subsubsection{Text-to-Text Generation}\label{subsubsect:text2text}

In the text-to-text generation category, we employed three advanced LLMs:  GPT4o (OpenAI's ChatGPT)~\cite{openai2023chatgpt, openai2024gpt4o}, GPT4 (Microsoft's Copilot)~\cite{microsoft2023copilot}, and Gemini-1.5-Flash-8B (Google AI Studio). We chose these LLMs based on their robust capabilities in natural language understanding and generation, as well as their ability to handle complex CSV documents, which was critical for generating few-shot synthetic accelerometer data.

As illustrated in Figure~\ref{fig:data_generation}(b), for generating the accelerometer data, we employed two prompt strategies: zero-shot and few-shot learning. In the zero-shot approach, the LLMs generate data based solely on the provided prompt, without any additional examples. In contrast, the few-shot approach involves supplying the models with examples of the desired output, which improves its ability to produce more accurate and contextually relevant data. Specifically, we used data from five subjects in the baseline datasets in CSV documents, each experiencing different types of falls (back, front, left, right, and rotate), as shots in the few-shot approach. The system prompts for both strategies are shown in Figure~\ref{fig:data_generation}(b).

\subsection{Baseline Preparation}\label{subsubsect:data_prep}

\subsubsection{Basline Datasets}
In this study, we utilized four datasets as baselines: SmartFallMM, KFall, UMAFall, and SisFall. The details of these datasets are summarized in Table~\ref{tab:datasets_summary}.
\begin{table*}[ht]
    \centering
    \caption{Summary of four baseline datasets considering the number of participants, their ages and genders (M: Males, and F: Females), number of falls and ADLs, sensors used, and sampling rate.}
    \resizebox{\textwidth}{!}{ 
    \begin{tabular}{|l|c|c|c|c|c|c|c|}
        \hline
        \textbf{Dataset} & \textbf{Participants} & \textbf{Age (Years)} & \textbf{Gender (M/F)} & \textbf{Falls} & \textbf{ADLs} & \textbf{Sensors Used} & \textbf{\makecell{Sampling Rate\\ (Hz)}} \\
        \hline
        SMM & 42 (16 young, 26 old) & \makecell{23 (young), \\ 65.5 (old)} & \makecell{11M, 5F (young);\\ 12M, 14F (old)} & 5 types & 9 types & \makecell{Huawei Smartwatch (left wrist),\\ Nexus Smartphone (right hip)} & 32 \\
        \hline
        KFall & 32 (all young) & 24.9 (young) & 32M, 0F & 15 types & 21 types & LPMS-B2 (lower back) & 100 \\
        \hline
        UMAFall & 17 (all young) & 19–28 (young) & 10M, 7F & 3 types & 8 types & \makecell{SimpleLink SensorTag \\(waist, right wrist)} & 20 \\
        \hline
        SisFall & 38 (23 young, 15 old) & \makecell{19–30 (young),\\ 60–75 (old)} & \makecell{11M, 12F (young); \\ 8Males, 7F (old)} & 15 types & 19 types & ADXL345 (waist) & 200 \\
        \hline
    \end{tabular}
    }
    \label{tab:datasets_summary}
\end{table*}

The SmartFallMM (SMM) dataset~\footnote{Dataset can be downloaded from here: \url{https://anonymous.4open.science/r/smartfallmm-4588}} includes data from 16 young (11 male and 5 female, with an average age of 23) and 26 old (12 males and 14 females, with an average age of 65.5) participants. Only young participants were instructed to simulate 5 types of falls (front, back, left, right, and rotational) on an air mattress, in addition to performing 9 types of Activities of Daily Living (ADLs) performed by both young and old participants, with each activity repeated five times. The accelerometer data of falls and ADLs were collected using four types of sensors at a frequency of 32 Hz: Meta Sensors (from MBIENTLAB) on the right wrist and left hip, a Huawei Smartwatch on the left wrist, and a Nexus smartphone on the right hip. For this study, we used accelerometer data collected from the Huawei Smartwatch, worn on the left wrist, and the Nexus smartphone placed on the right hip as baselines.

The KFall~\cite{KFall} dataset includes simulations from 32 young participants (all males with an average age of 24.9) who performed 21 daily activities and 15 simulated falls, with each activity repeated 5 times. The dataset includes 21 types of ADLs and 15 types of simulated falls (from walking, sitting, fainting, trying to get up or sit down).
Data collection utilized a custom-designed system that synchronizes sensor data with high-frequency video. A nine-axis inertial sensor (LPMS-B2) was attached to the lower back of participants, recording data from a three-axis accelerometer, gyroscope, and magnetometer at a frequency of 100 Hz. For this study, only the accelerometer data was used.

The UMAFall~\cite{umafall2017} dataset comprises data collected from 17 young participants (10 males and 7 females aged between 19–28 years), with sensors placed at five body locations: ankle, waist, right wrist, chest, and right trouser pocket. The dataset includes 8 types of ADLs and 3 types of falls (backward, forward, and lateral falls). In this study, we used accelerometer data of the waist and right wrist only, collected from SimpleLink SensorTag units, at a sampling rate of 20 Hz.

The SisFall~\cite{sisfall2017} dataset comprises data from 38 participants, divided into two groups: 23 young adults (11 males and 12 females aged 19–30 years) and 15 elderly individuals (8 males and 7 females aged 60–75 years). The dataset includes 19 types of ADLs and 15 types of falls (falls from walking, fainting, jogging, and trying to get up or sit down. Sensors were placed at the waist of participants and secured using a belt buckle. Data collection was performed using a self-developed embedded device, which included a Kinets MKL25Z128VLK4 microcontroller, an Analog Devices ADXL345 accelerometer, a Freescale MMA8451Q accelerometer, and an ITG3200 gyroscope. In this study, we used data captured from the ADXL345 accelerometer, at a sampling rate of 200 Hz.

\subsubsection{Diffusion-generated Dataset}
To compare the performance of synthetic data generated from text-to-motion and text-to-text models with diffusion-generated data, we utilize Diffusion-TS~\cite{diffusionts2024}, a cutting-edge generative model specifically designed for time-series data. Diffusion-TS has been extensively validated on various time-series datasets, demonstrating superior data similarity, temporal consistency, and adaptability across different domains. For this study, it serves as a robust benchmark for evaluating the quality and impact of synthetic data in fall detection tasks. We trained the Diffusion-TS model for each dataset, and generated 2001 samples, giving us 4 sets of diffusion-generated data for SMM, KFall, UMAFall, and SisFall. 

\subsubsection{Data Preparation for Training and Testing a Fall Detection Model}
To establish baselines, we selected only 12 subjects from all datasets considering the computationally limited resources, and consistency. For training a fall detection model, we grouped the documents (all activities including falls and ADLs of particular subjects) based on subjects using a Leave-Two-Out cross-validation strategy. Specifically, from 12 subjects, data from 8 subjects were used for training, data from 2 subjects were used for validation during training, and data from the remaining 2 subjects was reserved for the final evaluation of the model. This method ensures that the performance of the model is assessed on completely unseen data, which provides a robust evaluation of its generalizability. Then, in each group, we processed the documents to generate the sliding windows, labeling the windows from ADL documents as 0, and those from fall documents as 1.

To generate synthetic data from Diffusion-TS, we trained it using only fall data of 12 subjects, individually from each baseline dataset, keeping 80\% for training and 20\% for testing the generated data. For training and testing the fall classifier, we incorporated synthetic data into the training set of each baseline dataset. For instance, merging the SMM dataset with T2M-generated synthetic data created one augmented dataset. Similarly, combining SMM with other text-to-motion and text-to-text models resulted in a total of thirteen distinct augmented datasets. We followed a data-distribution ratio of 60\% ADLs, 20\% real falls, and 20\% synthetic falls.

To evaluate the classifier, we trained the LSTM model separately for each combination of real and synthetic datasets. For example, for SMM, we trained the model using synthetic data from Diffusion-TS, three text-to-motion models, and nine text-to-text models, each added individually to the baseline dataset. In total, we trained 13 models for SMM, excluding the model trained solely on the baseline dataset. The same approach was applied to the other three baseline datasets.

To create sliding windows, we used an overlapping-window technique with a fixed window size of $W = 128$, as our previous work demonstrated its effectiveness in optimizing accuracy~\cite{s24196235}. The stride, or step size, was set to 10, meaning that each new window introduced 10 new data points while overlapping with the previous window by 118 data points. This way, we obtained samples of size $128 \times 3$, where each of the $128$ data points consisted of three coordinates representing the $x$, $y$, and $z$ axes of the accelerometer data.

The input samples from each dataset can be represented as follows: Consider an input sequence $\mathbf{S}_i \in \mathbb{R}^{W \times 3}$, where $\mathbf{S}_i$ denotes the input sample from the $i$-th dataset (out of 4 baseline and the 13 synthetic datasets), and each row $(x_j, y_j, z_j)$ corresponds to the $x$, $y$, and $z$ coordinates of the accelerometer at time step $j$, with $j = 1, 2, \dots, W$:
\begin{equation} 
    \mathbf{S}_i = \{(x_j, y_j, z_j) \mid j = 1, 2, \dots, W\}
\end{equation}\label{eq:input_sample}
where $\mathbf{S}_i$ represents the input sample from the $i$-th dataset, $W$ denotes the number of time steps which is set to 128, and $(x_j, y_j, z_j)$ are the accelerometer coordinates at each time step $j$.

\subsection{Fall Detection Classifier}\label{subsect:lstm}
To train/test a fall detection classifier, and assess the impact of synthetic data on fall detection performance, we use the Long Short-Term Memory (LSTM)-based neural network~\cite{hochreiter1997long}. 
LSTM is designed to capture temporal dependencies in sequential data, making it well-suited for analyzing time series such as accelerometer data~\cite{ngu2024empirical, Wu2021, Butt2022, Zhou2023, s24196235}. The model begins with an LSTM layer that consists of 128 units. This layer is followed by a Dense layer with 128 units and Rectified Linear Unit (ReLU) activation~\cite{agarap2018deep}. A BatchNormalization (BN) layer is applied to normalize the output. The final Dense layer, with a sigmoid activation function, outputs the prediction. Mathematically, the prediction $\hat{y}$ of model can be described as:
\begin{equation}
    \hat{y} = \sigma\left(w_2 \cdot \text{BN}\left(\text{ReLU}\left(w_1 \cdot \text{LSTM}(\mathbf{S}) + b_1\right)\right) + b_2\right)
\end{equation}

where $\sigma$ is the sigmoid function, $w_1$ and $w_2$ are weight matrices, $b_1$ and $b_2$ are bias terms, BN represents batch normalization, and $S$ represents the input sample prepared in previous step.

Since model architectures inherently influence fall detection performance, one might assume that selecting a more advanced model could improve results. However, our objective is not to optimize classification performance through model selection but to isolate the effect of synthetic data. We intentionally choose LSTM, a model that does not inherently achieve the highest accuracy to ensure that any observed performance changes are driven by the quality and alignment of synthetic data rather than architectural advantages or parameter tuning. This allows for a more controlled and unbiased evaluation of synthetic data integration. 
\section{Implementation Details and Evaluation Criteria}\label{sect:implement_detaisl}
For training the LSTM model, we standardized the input datasets by removing the mean (to maintain numerical stability in activations) and scaling to unit variance to ensure consistent gradient behavior and improve the training efficiency. We developed the model using the TensorFlow framework, compiled it with binary cross-entropy loss, and optimized it with the Adam optimizer. Early stopping was employed to monitor the training process, with a patience of 50 epochs and a maximum of 250 epochs. Shuffling was set to \emph{True} to introduce randomness in the training process, preventing the model from learning unintended patterns based on data order. A learning rate of 0.001 was consistently applied across all models. The training was conducted on a Dell Precision 7820 Tower, equipped with 256 GB of RAM and a GeForce GTX 1080 GPU.

To ensure robustness and mitigate overfitting, we conducted five iterations of training and testing, where both the subject selection and synthetic data sampling were randomized in each iteration. Specifically, in each iteration, we randomly selected 8 subjects for training, 2 for validation, and 2 for testing. Additionally, 20\% of synthetic samples were randomly chosen and incorporated into the training set. 

In this study, our primary focus is on the positive class labeled as 1, which represents falls. Therefore, we used the F1-score as the primary metric for evaluating and comparing the performance of the model, as it effectively measures the balance between precision and recall, particularly in the context of imbalanced class distributions. The F1-score is defined as the harmonic mean of precision and recall:
\begin{equation}
\text{F1-score} = 2 \times \frac{\text{Precision} \times \text{Recall}}{\text{Precision} + \text{Recall}}
\end{equation}
where Precision is the ratio of true positive predictions to the total predicted positives, and Recall is the ratio of true positive predictions to the total actual positives. 
The F1-score ranges from 0 to 1, a higher F1-score (closer to 1) for class 1 indicates better model performance in detecting falls, achieving a strong balance between precision and recall, while a lower F1-score (closer to 0) suggests that the model struggles either with false positives (low precision) or missed detections (low recall).

\section{Results and Discussion}\label{sect:results_disc}
\begin{figure}[tbh!]
  \centering
  \includegraphics[width=0.8\textwidth]{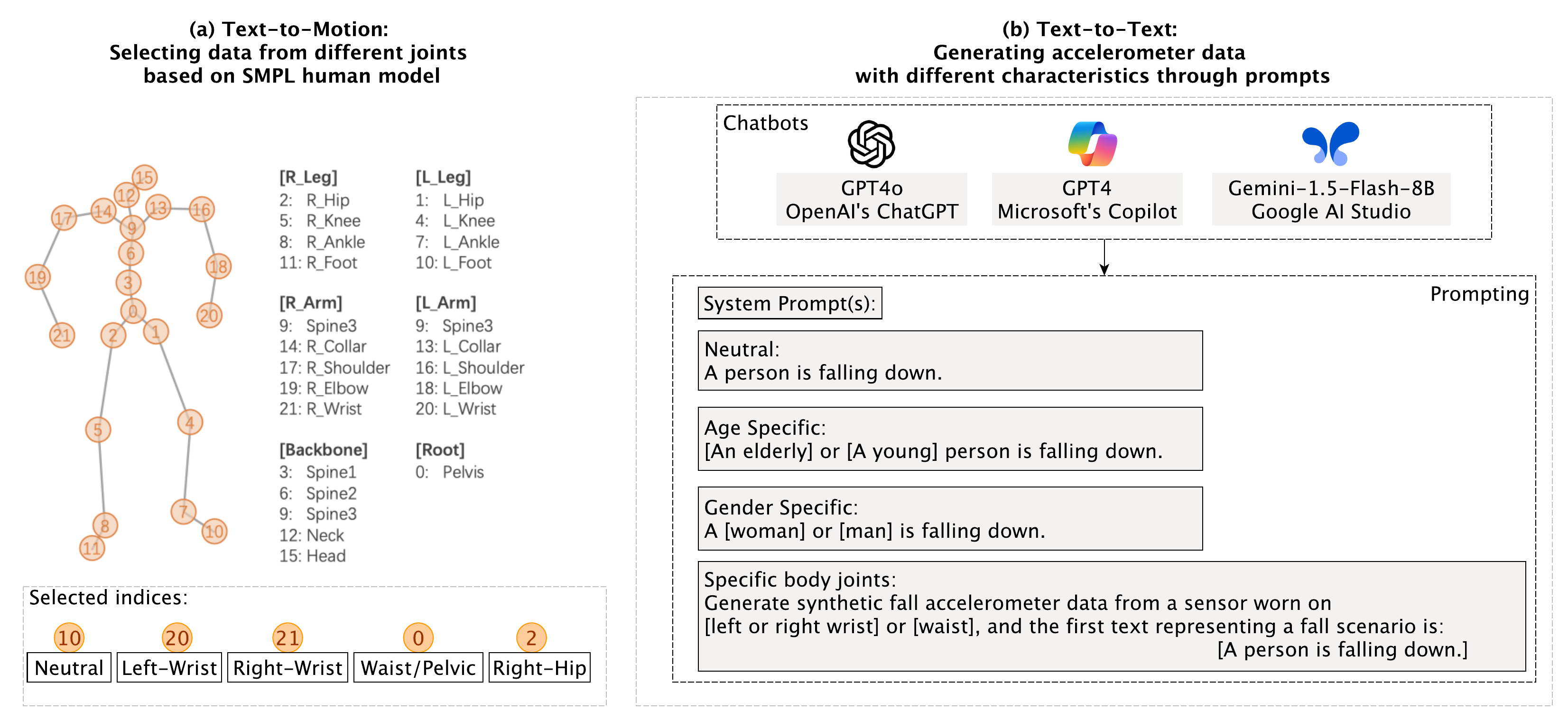}
  \caption{Process for generating synthetic accelerometer data with different characteristics using Text-to-Motion and Text-to-Text LLMs. (a) Text-to-Motion: Data selection from specific joints based on the SMPL human model, highlighting selected indices for neutral, left wrist, right wrist, and waist/pelvic data. (b) Text-to-Text: Prompt-based fall data generation using GPT4o, GPT4, and Gemini (1.5-Flash-8B).}
  \label{fig:selecting_joint_data}
\end{figure}
\subsection{Can LLMs generate accelerometer data specific to gender, age, and joint?}
To determine whether LLMs can generate accelerometer data specific to demographic and sensor placement characteristics, we conducted an experiment comparing neutral and specific prompt conditions. 
First, we created a set of neutral prompts that described generic fall scenarios without specifying demographic attributes or sensor locations. These prompts avoided references to gender, age, and body joints. We then designed additional prompt groups that explicitly incorporated these characteristics: gender-specific prompts (differentiating between men and women), age-specific prompts (distinguishing between young and elderly subjects), and joint-specific prompts (assigning sensor placements such as the left wrist, right wrist, and waist). In total, we obtained $50 \times 7 = 350$ prompts after permutations.

To extract joint-specific accelerometer data from text-to-motion LLMs, we utilized a 22-joint body system based on the widely used Skinned Multi-Person Linear (SMPL) model~\cite{loper2023smpl}. As illustrated in Figure~\ref{fig:selecting_joint_data}(a), each joint in the SMPL system is represented by a unique index, allowing precise selection of accelerometer data corresponding to specific body locations. Specifically, we selected the relevant indices as follows: the left wrist data was extracted from index 20, the right wrist from index 21, and the waist/pelvic region from index 0. For the neutral case, we deliberately chose the left foot, extracted from index 10. This selection was made to ensure that the reference data was biomechanically distinct from the upper-body sensor placements while still capturing meaningful movement patterns. The foot serves as a valid neutral baseline for two key reasons. First, during falls, lower extremity movements are often less affected by arm-specific actions such as reaching or bracing, making foot data less likely to encode gender- or age-specific motion variations. Second, the left foot provides a stable reference point that does not introduce bias toward common fall recovery mechanisms, which often involve arm or torso movements, thereby minimizing the risk of subtle demographic influences affecting our comparative analysis.

We computed joint-specific accelerometer data from the extracted 3D joint coordinates by calculating discrete second-order differences across consecutive frames. For example, considering the left wrist joint, let $\mathbf{p}(f) = [x(f), y(f), z(f)]$ denote 3D positional coordinates at frame $f$. We compute the acceleration vector $\mathbf{a}(f)$ as the discrete second derivative of position with respect to time as follows:
\begin{equation}
\mathbf{a}(f) = \frac{\mathbf{p}(f+1) - \mathbf{p}(f)}{\Delta t^2}
\end{equation}
where $\Delta t$ represents the time interval between frames ($\Delta t = 1/46\,\text{s}$). We independently computed each component of acceleration, namely $a_x(f)$, $a_y(f)$, and $a_z(f)$, from positional differences along their corresponding axes, scaling appropriately to obtain units in meters per second squared ($\text{m/s}^2$). 

For text-to-text models, we generated synthetic accelerometer data by directly prompting the models with different specificity levels. This process is illustrated in Figure~\ref{fig:selecting_joint_data}(b). The LLMs were instructed to generate biomechanical time-series outputs based on zero-shot prompts that either included demographic details (e.g., age and gender) or specified joint locations. 

\begin{figure}[tbh!]
  \centering
  \includegraphics[width=1.0\textwidth]{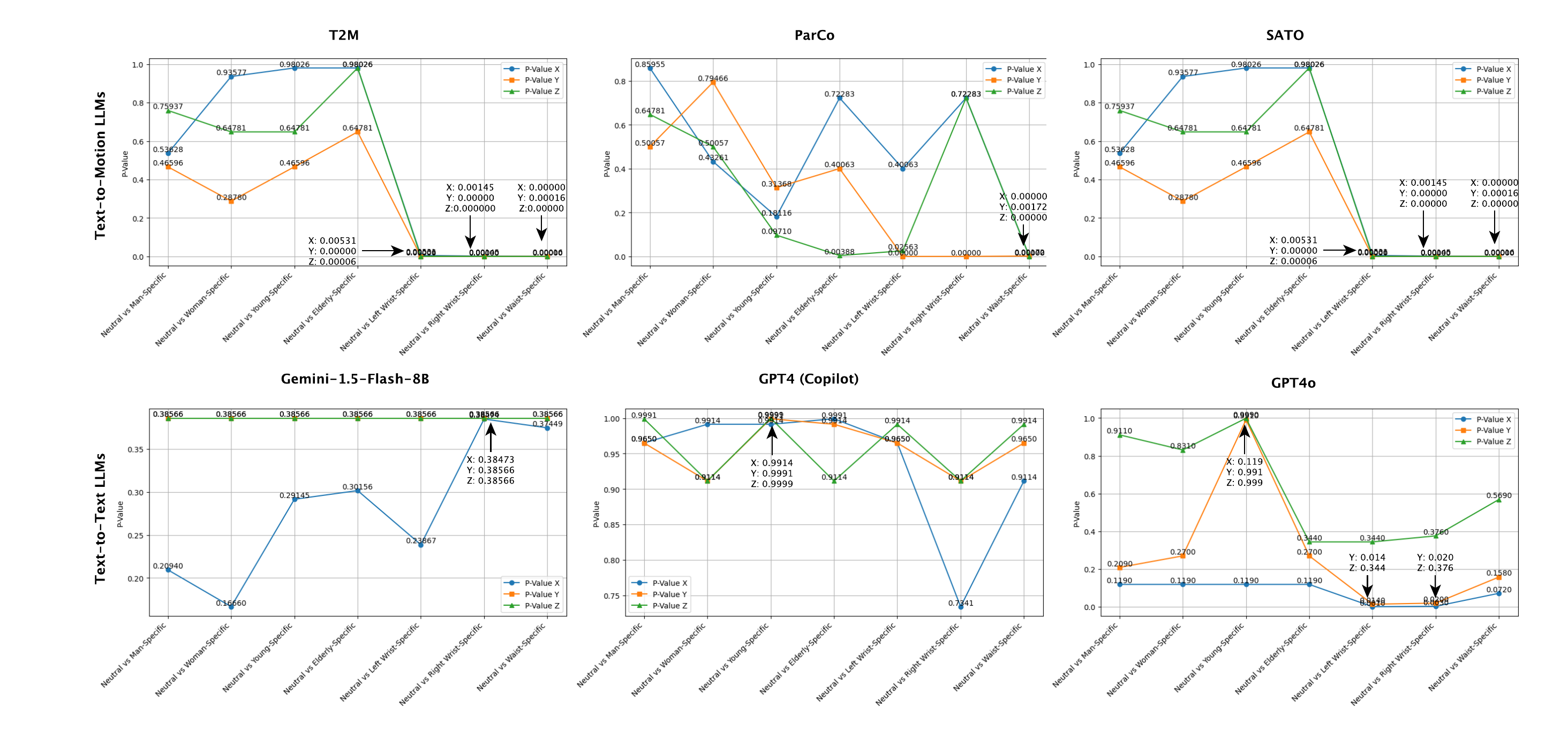}
  \caption{P-values from Kolmogorov-Smirnov tests comparing neutral and specific datasets for Text-to-Motion (T2M, ParCo, and SATO) and Text-to-Text LLMs (Gemini, GPT4, and GPT40).}
  \label{fig:lllm_ktest_pvalues}
\end{figure}
After collecting the synthetic accelerometer data from the models of both groups, we employed the Kolmogorov-Smirnov (KS) test~\cite{kolmogorov1933}, a non-parametric method for comparing the distribution of two independent datasets. This test calculates KS statistics and corresponding p-values, which quantify the likelihood that the two datasets originate from the same distribution. A p-value closer to 1 (above 0.05) suggests that the two datasets are statistically similar, indicating that the LLM did not significantly alter its generated accelerometer data based on the prompt variations. Conversely, a p-value close to 0 (below 0.05) suggests that the datasets differ significantly, implying that the LLM incorporated the specified demographic or joint-specific characteristics in a meaningful way.

The KS test results for text-to-motion models, T2M, ParCo, and SATO, reveal distinct trends in how accelerometer data distributions vary across demographic and joint-specific conditions. As shown in Figure~\ref{fig:lllm_ktest_pvalues} (top row), for gender and age-specific comparisons (Neutral vs Man, Woman, Young, and Elderly), the p-values remain relatively high across all axes ($X$, $Y$, and $Z$). Specifically, the average p-values for T2M range from 0.64781 to 0.98026, for SATO from 0.64781 to 0.98026, and for ParCo from 0.19731 to 0.66931, all of which are above the 0.05 threshold, indicating no statistically significant differences between the neutral dataset and those generated for men, women, young, or elderly individuals.  
However, when comparing neutral data with joint-specific datasets, particularly for left wrist, right wrist, and waist placements, the p-values drop significantly, often approaching zero. 
For the left wrist, the average p-values are 0.00179 for T2M, 0.00179 for SATO, and 0.14209 for ParCo, with ParCo showing a slightly weaker distinction compared to T2M and SATO. For the right wrist, the values are 0.00081 for T2M, 0.00081 for SATO, and 0.48189 for ParCo, with ParCo displaying much higher values, indicating less distinction from the neutral data. For the waist, the values are 0.00005 for T2M, 0.00005 for SATO, and 0.00057 for ParCo, all of which confirm strong differences from the neutral dataset.
The most pronounced differences are observed in the $Y$ and $Z$ comparisons for wrist and waist placements, as the p-values are below the 0.05 threshold, implying that the model encodes different movement dynamics depending on the specified joint. Notably, the highest variance is observed in T2M and SATO, where the p-values for left wrist, right wrist, and waist remain consistently low across all axes, while ParCo exhibits comparatively higher p-values, particularly for right wrist placement, suggesting less divergence from the neutral dataset. 

For text-to-text models, the average p-values for demographic-specific comparisons are 0.685 for GPT-4o, 0.733 for GPT-4, and 0.707 for Gemini, all of which are well above the 0.05 threshold, confirming that these models do not generate significantly different distributions for different demographic categories. GPT-4o exhibits the most variability across conditions, with lower p-values for joint-specific prompts, particularly for the left and right wrist, suggesting that it differentiates motion patterns based on sensor placement more than the other models. Its average p-value for joint-specific comparisons is 0.00267, indicating a strong sensitivity to sensor placement variations. However, its demographic-specific data, such as gender and age, show higher p-values, indicating little statistical distinction from neutral data.
In contrast, GPT-4 (Copilot) consistently produces high p-values across all comparisons, with an average of 0.733 for demographic and 0.049 for joint-specific data, suggesting that its generated accelerometer data remain largely invariant regardless of demographic or sensor-specific conditions. This indicates that GPT-4 (Copilot) lacks the sensitivity to generate nuanced biomechanical variations based on prompt modifications.
Meanwhile, Gemini presents intermediate results, with an average p-value of 0.707 for demographic conditions and 0.0257 for joint-specific data, implying that it recognizes some level of sensor placement differences but is less pronounced than GPT-4o. Notably, the p-values of Gemini, for left and right wrist placements, are slightly higher than GPT-4o, indicating a weaker ability to differentiate joint-specific variations.
Based on these findings, GPT-4o appears to be the most responsive to joint-specific modifications, with noticeable changes primarily in the $Y$ and $Z$ coordinate values. However, these changes are mild, and the p-values remain significantly larger than those observed in text-to-motion models, indicating a weaker differentiation in biomechanical variations.

\begin{table}[htb!]
    \centering
    \caption{List of Prompts Used for Synthetic Data Generation}\label{appendix:prompts}
    \resizebox{0.75\textwidth}{!}
    {
    \begin{tabular}{|c|l|}
        \hline
        \textbf{Index} & \textbf{Prompts} \\ \hline
        1 & An elderly person is falling down. \\ \hline
        2 & An elderly person sits, becomes unconscious, then falls on his left, and lies on the floor. \\ \hline
        3 & An elderly person walks, becomes unconscious, then falls on his head, and lies on the floor. \\ \hline
        4 & An elderly person walks, then he falls down on his face, and stays on the floor. \\ \hline
        5 & An elderly person falls down on his right side, and lies on the ground. \\ \hline
        6 & A person stands still, wobbles, leans right suddenly, then falls down, and tucks in tight on the floor. \\ \hline
        7 & A person walks, slips suddenly, and falls on his back on the floor. \\ \hline
        8 & A person climbs a ladder, loses grip on one rung, and falls sideways. \\ \hline
        9 & A person walks on gravel, twists their ankle, and falls forward onto their hands and knees down to the floor. \\ \hline
        10 & A person runs, then slips and falls flat on the floor. \\ \hline
        11 & A person walks, suddenly trips, loses his balance, and ends up lying on the floor. \\ \hline
        12 & A person runs, slips, and FALLS FLAT on the floor. \\ \hline
        13 & A person walks, suddenly trips, loses his balance, and falls down on the floor, unable to get up. \\ \hline
        14 & A person walks, suddenly trips, loses his balance, and falls down and then ends up lying on the floor. \\ \hline
        15 & A person missed a chair and fell on the floor on his head. \\ \hline
        16 & A person falls asleep, falls down on the floor on his head. \\ \hline
        17 & A person is standing up, and falls down on the floor on his head. \\ \hline
        18 & A person falls down on the floor on his head. \\ \hline
        19 & A person walks, falls asleep, and then falls down on the floor on his head. \\ \hline
        20 & A child rides a bike too fast, loses control, and tumbles onto the grass with a scraped knee. \\ \hline
        21 & A person is stepping up, and falls down on the floor on his head. \\ \hline
        22 & A person walks, and then falls down on his head on the floor. \\ \hline
        23 & A person stands up from a chair, and falls down on the floor on his head. \\ \hline
        24 & A person bends to pick an object, and falls down on the floor on his head. \\ \hline
        25 & A person climbs stairs and falls down on the floor on his head. \\ \hline
        26 & A person kneels down to pick an object, slips, and falls on his head on the floor. \\ \hline
        27 & A person missed a chair and fell on the floor on his left. \\ \hline
        28 & A person missed a chair and fell on the floor on his back. \\ \hline
        29 & A person bends to the left, picks up something, becomes unconscious, and falls down onthe  floor on his head. \\ \hline
        30 & A person turns right, and then falls down on the floor on his head. \\ \hline
        31 & A person walks forward, becomes unconscious, and falls down on the floor on his head. \\ \hline
        32 & A person is walking. They suddenly trip, lose their balance, and end up lying on the floor. \\ \hline
        33 & A person walking suddenly trips, loses their balance, and ends up lying on the floor, unable to get up. \\ \hline
        34 & A person walking suddenly trips, loses their balance, and ends up lying on the floor. \\ \hline
        35 & A person is standing upright, then suddenly trips and ends up lying in the fetal position. \\ \hline
        36 & A person stands up. Then they descend to the floor, lies sprawled out. \\ \hline
        37 & A person attempts to get out of their chair, but ends up hitting their head on the floor and lying sprawled out. \\ \hline
        38 & \makecell[l]{A person is standing upright. They do a half turn before they lose their balance and descend to the floor.\\ They lie with their legs pulled up to their chest.} \\ \hline
        39 & A person is taking confident strides, until they trip hitting their head as they lie on the ground sprawled out. \\ \hline
        40 & A person walks forward and trips ending on the floor. \\ \hline
        41 & A person walks backwards and trips and falls ending on the floor. \\ \hline
        42 & A person walks on a wet floor, slips, then falls and lands on their side. \\ \hline
        43 & An elderly person stumbles while stepping off a curb, falls, and lands on their hip. \\ \hline
        44 & A person reaches for an object on a high shelf, loses balance, and falls backward on the floor. \\ \hline
        45 & A person runs down a hill, loses control, and rolls to a stop on the ground. \\ \hline
        46 & A person stands on a chair to change a light bulb, the chair tips, and they fall to the floor. \\ \hline
        47 & A person trips over his own feet, falls on the ground, and ends up lying there. \\ \hline
        48 & A person rides a scooter, hits a bump, and falls, landing on their knee. \\ \hline
        49 & A person leans too far back in a chair, and he falls backward on the floor. \\ \hline
        50 & A person walks downstairs, missteps, and falls down on the floor on his back. \\ \hline
    \end{tabular}
    }
    \label{tab:prompts}
\end{table}
\noindent\textbf{Findings:} The findings demonstrate that text-to-motion models, T2M, ParCo, and SATO, are highly effective at capturing joint-specific variations, making them ideal for applications requiring realistic and detailed motion patterns tied to specific sensor placements, such as wearable sensor training for fall detection. However, these models show limited differentiation for demographic attributes like gender and age. Among text-to-text models, GPT4o emerges as the most responsive, displaying some ability to generate joint-specific variations, though these changes are milder than those produced by text-to-motion models, making it suitable for scenarios where moderate differentiation suffices. 

Based on these findings, we designed our final set of prompts without specifying sensor locations, also ensuring that the majority are devoid of demographic attributes. These prompts were used to generate synthetic data, which serves as the foundation for our research presented in later sections. The complete list of prompts is provided in Table~\ref{tab:prompts}.

\subsection{Do LLM-generated data perform better or align more closely with real data than diffusion-based methods?}
\label{subsect:align_syn_vs_diff}
To evaluate the alignment of synthetic fall data with real fall datasets, we conducted two types of analyses: qualitative analysis using visualizations and quantitative analysis using two performance metrics. 
To prepare comparison pairs for each baseline dataset, we utilized zero-shot (ZS) synthetic data generated by text-to-text models without modification. For example, when comparing baseline SMM data with text-to-text model outputs, we directly used ZS data from each model. In contrast, to generate few-shot (FS) data, we provided each model with examples of falls from five randomly selected subjects per dataset. Specifically, for the SMM, KFall, UMAFall, and SisFall datasets, we extracted fall samples from five subjects and submitted them as CSV files, resulting in 12 ($4 \times 3$) sets of FS accelerometer data across the three text-to-text models.
For text-to-motion generation, we extracted data from the relevant joint indices corresponding to each baseline dataset. For instance, for the SMM dataset, we used data from the left wrist (index 20), for KFall and SisFall, the waist (index 0), and for UMAFall, the right wrist (index 21) data (see Figure~\ref{fig:selecting_joint_data}).

For both qualitative and quantitative analysis, we generated 28 synthetic datasets for comparison with four real-world datasets. These included 4 datasets from the Diffusion-TS model, 12 from text-to-text models, and 12 from text-to-motion models.

\begin{figure}[tbh!]
  \centering
  \includegraphics[width=1.0\textwidth]{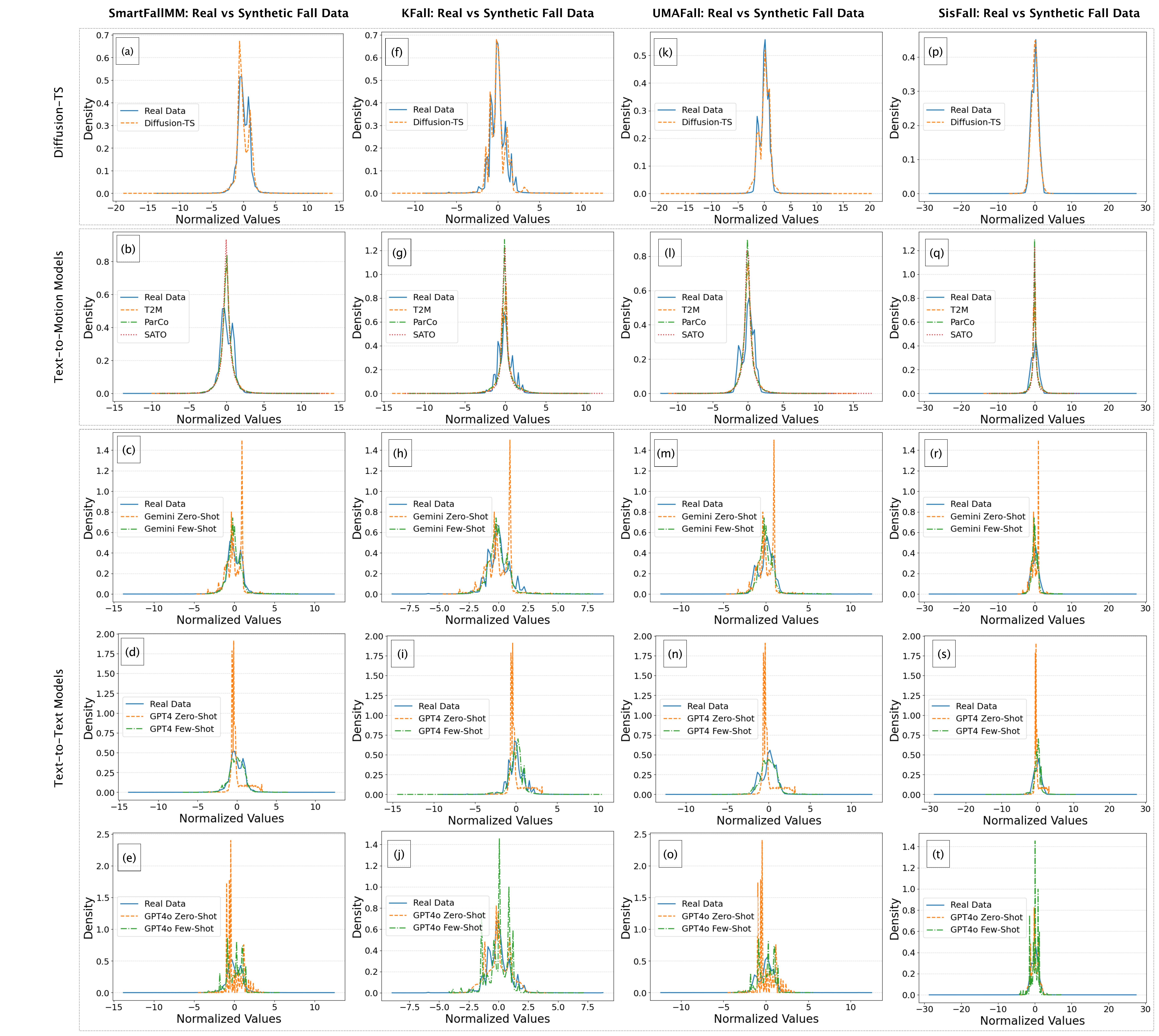}
  \caption{Comparison of normalized value distributions of real falls of four datasets, SmartFallMM, KFall, UMAFall, and SisFall, versus synthetic fall data generated from diffusion-based, text-to-motion, and text-to-text models.}
  \label{fig:llm_synthetic_visualization}
\end{figure}
\noindent \textbf{Qualitative Analysis:} In this analysis, we employed line plots to compare the distribution of synthetic falls generated by diffusion-based, text-to-motion, and text-to-text models with real falls from the SMM, KFall, UMAFall, and SisFall datasets. Figure~\ref{fig:llm_synthetic_visualization} illustrates these comparisons. The line plots, derived from histogram data, plot bin centers against histogram density values to create a smooth approximation of the distribution. Distinct line styles are used for each dataset to enhance clarity.

\textit{SMM Dataset:} In Figure~\ref{fig:llm_synthetic_visualization}(a), Diffusion-TS closely aligns with real data but exhibits minor discrepancies, such as subtle peaks and valleys that deviate slightly from the real distribution. In Figure~\ref{fig:llm_synthetic_visualization}(b), text-to-motion models vary in performance, with SATO showing the largest peak, indicating over-concentration at the mean. ParCo also displays a pronounced peak, though less extreme than SATO, while T2M aligns best with real data, exhibiting only slight overestimation and maintaining good variability. 
Among text-to-text models, Gemini-ZS (Figure~\ref{fig:llm_synthetic_visualization}(c)) produces a sharp peak at the center, reflecting poor generalization and limited variability. Gemini-FS improves alignment with a lower peak and more realistic spread. Similarly, in Figure~\ref{fig:llm_synthetic_visualization}(d), GPT4-ZS exhibits the most extreme over-concentration near the mean, while GPT4-FS captures greater variability but still deviates from real data. GPT4o-ZS (Figure~\ref{fig:llm_synthetic_visualization}(e)) follows the same pattern, with a large peak and reduced variance, whereas GPT4o-FS aligns better but still overestimates density near the mean. Across these models, ZS models perform the worst, showing excessive central density and minimal variability. The Few-Shot models provide some improvement, however, alignment issues remain. 
The results suggest that, for the SMM dataset, text-to-motion models outperform text-to-text models, with Diffusion-TS and T2M providing the closest approximations to real data.

\textit{KFall Dataset:} In Figure~\ref{fig:llm_synthetic_visualization}(f), Diffusion-TS demonstrates strong alignment with real data, effectively capturing the central tendency and variability, with only minor deviations in specific regions. In Figure~\ref{fig:llm_synthetic_visualization}(g), text-to-motion models vary in performance, with T2M showing the smallest peak and closely matching real data, effectively preserving both spread and variability. ParCo exhibits a slightly larger peak, indicating minor over-concentration near the mean, yet still maintains good distributional alignment. SATO, however, displays the largest peak among these models, significantly exceeding the density of real data near the mean, suggesting an over-concentration of highly probable fall patterns and reduced variability. 

Among text-to-text models, Gemini-ZS (Figure~\ref{fig:llm_synthetic_visualization}(h)) produces a large peak at the center, indicating poor generalization and a lack of variability. Gemini-FS improves alignment by reducing the peak and better capturing the spread, slightly tending to over-concentrate near the mean. Similarly, in Figure~\ref{fig:llm_synthetic_visualization}(i), GPT4-ZS exhibits the largest peak among the text-to-text models, heavily over-concentrating density at the mean and underestimating variability. GPT4-FS shows better alignment, reducing peak intensity and capturing greater variability with minimal deviations. In Figure~\ref{fig:llm_synthetic_visualization}(j), GPT4o-FS exhibits a reverse trend, performing worse than its Zero-Shot counterpart, resulting in spiky and less representative distributions. 

Across all models, for the KFall dataset, Diffusion-TS and T2M exhibit the closest alignment to real data, effectively capturing both central tendency and variability. While FS models generally improve alignment, certain cases, such as GPT4o-FS, demonstrate that additional training examples can sometimes degrade distributional accuracy, introducing noise and increased variability in the synthetic data.

\textit{UMAFall Dataset:} In Figure~\ref{fig:llm_synthetic_visualization}(k), Diffusion-TS closely aligns with real data, effectively capturing both the central tendency and spread of the distribution with minimal deviations. In Figure~\ref{fig:llm_synthetic_visualization}(l), text-to-motion models display distinct trends, with ParCo exhibiting the largest peak, significantly over-concentrating density at the mean while underestimating variability. SATO shows a medium-sized peak, slightly lower than ParCo, indicating moderate over-concentration near the mean but better variability. T2M demonstrates the smallest peak, offering the best alignment with real data by effectively capturing both central tendency and variability. 

Among text-to-text models, Gemini-ZS (Figure~\ref{fig:llm_synthetic_visualization}(m)) produces a sharp peak that significantly exceeds the density of real data at the mean, with limited variability, reflecting an over-concentration on central trends. Gemini-FS reduces the peak height and improves alignment, capturing more variability, however, the distribution appears noisier and lacks smoothness. Similarly, in Figure~\ref{fig:llm_synthetic_visualization}(n), GPT4-ZS exhibits the largest peak among text-to-text models, over-concentrating density near the mean, and underestimating variability, indicative of poor generalization. GPT4-FS reduces the peak height and improves alignment by capturing broader variability with minor deviations in the tails of the distribution. In Figure~\ref{fig:llm_synthetic_visualization}(o), GPT4o-ZS displays a sharp peak similar to GPT4-ZS, suggesting poor variability. 

Across all models for the UMAFall dataset, Diffusion-TS and T2M emerge as the most reliable in capturing real data distributions, effectively balancing central tendency and variability. 

\textit{SisFall Dataset:} In Figure~\ref{fig:llm_synthetic_visualization}(p), Diffusion-TS achieves near-perfect alignment with real data, effectively capturing both central tendency and variability with minimal deviations. Among text-to-motion models (Figure~\ref{fig:llm_synthetic_visualization}(q)), T2M demonstrates the best performance, exhibiting the smallest peak and closely matching the real data distribution. SATO shows a medium-sized peak, indicating slight over-concentration near the mean but maintaining reasonable variability. In contrast, ParCo displays the largest peak, significantly over-concentrating density at the mean, leading to poorer alignment. 
Text-to-text models reveal varied performance trends. Gemini-ZS (Figure~\ref{fig:llm_synthetic_visualization}(r)) produces a sharp, exaggerated peak at the center, reflecting poor generalization and minimal variability. Gemini Few-Shot reduces the peak height and captures broader variability with minimal noise. Similarly, GPT4-ZS (Figure~\ref{fig:llm_synthetic_visualization}(s)) exhibits the most pronounced over-concentration at the mean among text-to-text models, significantly underestimating variability. GPT4-FS improves upon this by reducing the peak height and broadening variability, with minimal deviations, particularly in the tails of the distribution. In Figure~\ref{fig:llm_synthetic_visualization}(t), GPT4o-FS exhibits the largest peak, indicating excessive concentration near the mean, with noticeable spikiness and noise, reducing overall alignment. In contrast, GPT4o-ZS maintains better alignment with the base distribution, capturing central tendency more effectively and displaying smoother variability. This reflects the reverse trend also observed in KFall, where Few-Shot configurations for this model performed worse than ZS. 
Across the models for the SisFall dataset, Diffusion-TS and T2M demonstrate the strongest alignment with real data. 

\noindent \textbf{Quantitative Analysis:}
In this section, we quantitatively evaluate the alignment of synthetic fall data with real datasets using two metrics: Coverage~\cite{naeem2020} and Jensen-Shannon Divergence (JSD)~\cite{endres2003new}. Coverage measures the proportion of real data samples represented by synthetic data. Higher values indicate effective capture of real fall variability, suggesting strong generalization. JSD quantifies the statistical similarity between real and synthetic distributions. Lower JSD values signify closer alignment, indicating high fidelity. 

\begin{figure}[tbh!]
  \centering
  \includegraphics[width=1\textwidth]{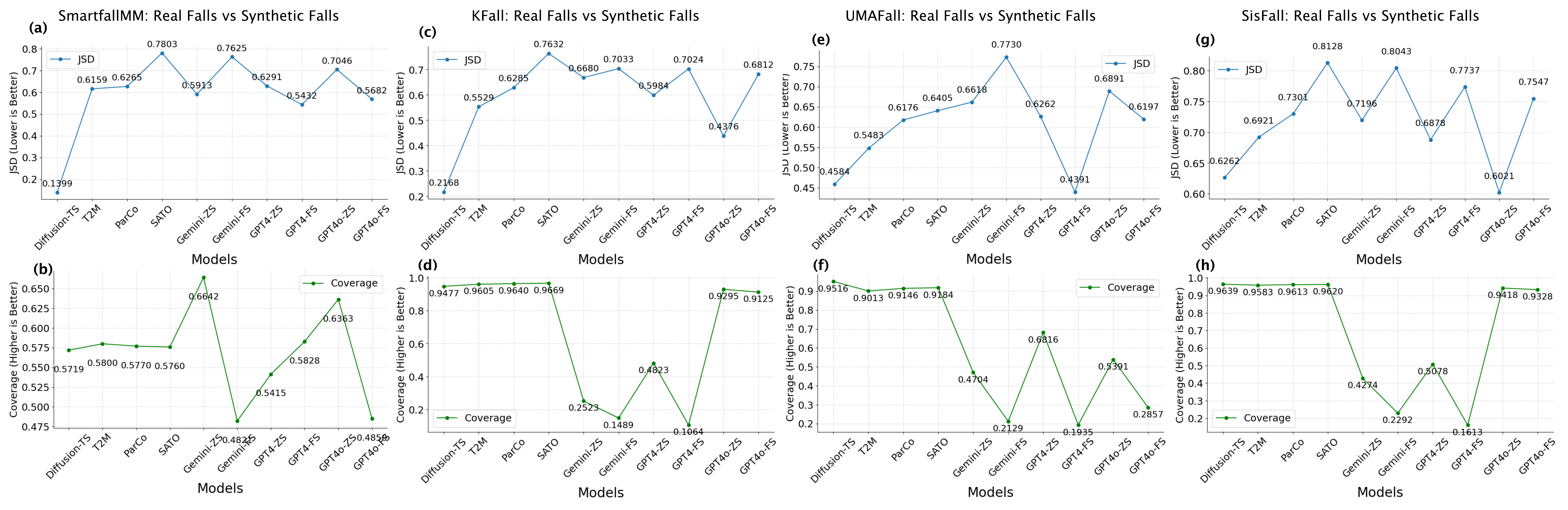}
  \caption{Quantitative comparison of real falls of four baseline datasets versus synthetic falls generated by different LLMs, evaluated in terms of Jensen-Shannon Divergence (JSD) and Coverage.}
  \label{fig:llms_quantitative_analysis}
\end{figure}
\textit{For the SMM dataset}, as illustrated in Figure~\ref{fig:llms_quantitative_analysis}(a) and (b), Diffusion-TS exhibits the lowest JSD (0.1399) and a moderate Coverage (0.5719) among all models, indicating the best alignment with real data. T2M, ParCo, and SATO show significantly higher JSD values (0.6159, 0.6265, and 0.7803, respectively), suggesting a greater divergence from real data. Coverage values for these models are comparable, ranging between 0.576 and 0.580, which are slightly higher than Diffusion-TS but do not compensate for the substantially higher JSD values. Among these three models, T2M achieves the lowest JSD and highest Coverage, aligning with the qualitative visualization where T2M exhibited fewer peaks and better distributional variability compared to ParCo and SATO. 

Among the Gemini models, Gemini-ZS achieves the highest Coverage (0.6642) but still has a relatively high JSD (0.5913), indicating that while it covers more data instances, it does not align well with the real distribution. Gemini-FS has a higher JSD (0.7625) and the lowest Coverage (0.4821) among all models, indicating weaker performance compared to Diffusion-TS. GPT4-ZS and GPT4-FS have JSD values of 0.6291 and 0.5432, respectively, suggesting moderate alignment with real data. Their Coverage scores (0.5415 and 0.5828) remain competitive, but they do not outperform Diffusion-TS overall. The GPT4o variants show a similar trend, with GPT4o-ZS achieving higher Coverage (0.6363) but a relatively high JSD (0.7046). GPT4o-FS performs better in terms of JSD (0.5682), though its Coverage (0.485) is lower than that of Diffusion-TS. 

\textit{In the KFall dataset}, as illustrated in Figure~\ref{fig:llms_quantitative_analysis}(c) and (d), Diffusion-TS again achieves the lowest JSD (0.2168) and the highest Coverage (0.9477), reinforcing its superior performance in capturing real data characteristics. T2M, ParCo, and SATO exhibit progressively increasing JSD values (0.5529, 0.6285, and 0.7632), with relatively high Coverage values exceeding 0.96 for T2M, ParCo, and SATO. Among these models, T2M achieves the lowest JSD (0.5529), indicating the best alignment with real data, as observed in the distribution plot (see Figure~\ref{fig:llm_synthetic_visualization}(g)). SATO has the highest Coverage (0.9669) likely due to its over-concentrated density near the mean (see Figure~\ref{fig:llm_synthetic_visualization}(g)), inflating its Coverage despite higher divergence. 

For text-to-text models, Gemini-ZS and Gemini-FS show significantly lower Coverage (0.2523 and 0.1489, respectively) and high JSD values (0.668 and 0.7033), indicating weaker performance. Among the GPT4 variants, GPT4o-ZS demonstrates the lowest JSD (0.4376) and high Coverage (0.9295), outperforming GPT4-ZS (0.5984 JSD, 0.4823 Coverage) and GPT4-FS (0.7024 JSD, 0.1064 Coverage). However, as observed in Figure~\ref{fig:llm_synthetic_visualization}(j), GPT4o-FS exhibited a reverse trend, performing worse than ZS due to increased variance and spikiness. This suggests that the higher JSDs of Gemini-FS, GPT4-FS, and GPT-4o-FS may stem from the irregularities and noise introduced.

\textit{For the UMAFall dataset}, as shown in Figure~\ref{fig:llms_quantitative_analysis}(e) and (f), Diffusion-TS maintains its leading performance with the lowest JSD (0.4584) and highest Coverage (0.9516). T2M, ParCo, and SATO have moderately higher JSD values (0.5483, 0.6176, and 0.6405, respectively), with Coverage values remaining above 0.9. Among these models, T2M achieved the lowest JSD (0.5483), and SATO achieved the highest coverage (0.9184) aligning with the qualitative analysis (see Figure~\ref{fig:llm_synthetic_visualization}(l)), where T2M exhibited the smallest peak and the best alignment with real data. 

Gemini-ZS and Gemini-FS exhibit poor performance, with JSD values exceeding 0.66 and much lower Coverage (0.4704 and 0.2129). GPT4-FS achieves the lowest JSD (0.4391) among them (even lower than Diffusion-TS), suggesting strong alignment. However, it has a very low Coverage (0.1935), indicating that while its distribution closely follows real data, it does not sufficiently capture the full range of variability. This is consistent with the distribution plot in Figure~\ref{fig:llm_synthetic_visualization}(n), where GPT4-FS demonstrated improved variability but still showed deviations in the tails. GPT4-ZS and GPT4o-ZS offer a balance between JSD and Coverage (0.6262 JSD, 0.6816 Coverage, and 0.6891 JSD, 0.5391 Coverage, respectively), while GPT4o-FS introduced additional noise and spikiness, hence producing higher JSD (0.6197) and lower Coverage (0.2857) than GPT4o-ZS.

\textit{For the SisFall dataset}, the Diffusion-TS model continues its trend of superior performance with the lowest JSD (0.6262) and the highest Coverage (0.9639), as shown in Figure~\ref{fig:llms_quantitative_analysis}(g) and (h). T2M, ParCo, and SATO follow with increasing JSD values (0.6921, 0.7301, and 0.8128), while maintaining relatively high Coverage (0.9583, 0.9613, and 0.9620, respectively). Among these models, T2M achieves the lowest JSD (0.6921), and ParCo provides the highest Coverage (0.9613). These statistics align with qualitative analysis (Figure~\ref{fig:llm_synthetic_visualization}(q)), where T2M exhibited the smallest peak and the closest alignment to real data, whereas SATO achieved highest Coverage, suggesting that, despite over-concentrating data near the mean, it captures more data instances but does not distribute them as effectively across the full range of variability.

For text-to-text models, Gemini-ZS and Gemini-FS again perform poorly, with Gemini-FS having the lowest Coverage (0.2292) and one of the highest JSD values (0.8043). 
Among GPT4 models, GPT-4-ZS achieves a moderate JSD (0.6878) and a Coverage of 0.5078, indicating better distributional alignment than Gemini models but still showing limitations in capturing the full variability of real data. GPT-4-FS, with a JSD of 0.7737 and the lowest Coverage (0.1613), struggles more with maintaining distributional coverage.
For GPT4o models, GPT-4o-ZS achieves the lowest JSD (0.6021) among all models (even lower than Diffusion-TS) and a relatively high Coverage (0.9418), suggesting a better balance between alignment and distributional coverage. GPT-4o Few-Shot, in contrast, has a higher JSD (0.7547) and slightly lower Coverage (0.9328), reflecting increased concentration around specific modes and potential overfitting to the few-shot examples (as observed in Figure~\ref{fig:llm_synthetic_visualization}(t)). 

\noindent\textbf{Findings:}
The qualitative and quantitative analyses indicate that Diffusion-TS consistently outperforms LLM-generated data in aligning with real data across multiple datasets. Diffusion-TS achieves the lowest JSD values and highest Coverage scores, effectively capturing both central tendency and variability with minimal deviations. In contrast, LLM-generated data, particularly from text-to-text models like GPT4 and GPT4o, exhibit inconsistent but slightly better performance than Gemini. The ZS configurations generally perform poorly, with sharp peaks, over-concentration near the mean, and reduced variability. Among LLMs, text-to-motion models like T2M outperform text-to-text models, showing smaller peaks and closer alignment with real data, yet they still fall short of the robustness of Diffusion-TS.

Additionally, the reverse trends observed in GPT4o FS for KFall and SisFall suggest that FS settings can introduce noise and spikiness, worsening alignment rather than improving it. Notably, these trends were only observed in datasets with higher sampling rates (Kfall: 100 Hz, and SisFall: 200 Hz), indicating that increased frequency amplifies the limitations of text-to-text models. As the sampling rate increased, text-to-text models exhibited a decline in performance, with higher JSD values, lower Coverage scores, and noisier distributions, pointing to a loss of specificity and accuracy. Meanwhile, Diffusion-TS remained robust across all baselines with different sampling rates, consistently maintaining strong alignment with real data.

\subsection{Do LLM-generated data improve the performance of the model for a fall detection task?}\label{susect:stats_synthetic}
\begin{table}[h!]
    \centering
    \caption{Performance comparison of various synthetic data generation methods for fall detection, showing percentage increase (\(+\)) or decrease (\(-\)) in the average F1-score of the LSTM model relative to the baseline. Results are reported across 5-fold cross-validation.}
    \resizebox{1\textwidth}{!}{ 
    \begin{tabular}{|l|c|c|c|c|c|c|c|c|c|c|c|}
        \hline
        \multirow{3}{*}{\textbf{Dataset}} & \multirow{3}{*}{\textbf{Baseline}} & \multirow{3}{*}{\textbf{Diffusion-TS}} & \multicolumn{3}{c|}{\textbf{Text-to-Motion LLMs}} & \multicolumn{6}{c|}{\textbf{Text-to-Text LLMs}} \\ \cline{4-12} 
                         &                   &                       & \multirow{2}{*}{\textbf{T2M}} & \multirow{2}{*}{\textbf{ParCo}} & \multirow{2}{*}{\textbf{SATO}} & \multicolumn{2}{c|}{\textbf{Gemini-Flash-8B}} & \multicolumn{2}{c|}{\textbf{GPT4 (Copilot)}} & \multicolumn{2}{c|}{\textbf{GPT4o (ChatGPT)}} \\ \cline{7-12} 
                         &                   &                       &              &                &               & \textbf{Zeroshot} & \textbf{Fewshot} & \textbf{Zeroshot} & \textbf{Fewshot} & \textbf{Zeroshot} & \textbf{Fewshot} \\ \hline
        SMM      & 0.740               & 0.680 (-8.11\%)        & \textbf{0.710 (-4.05\%)}  & 0.680 (-8.11\%)  & 0.648 (-12.43\%) & 0.636 (-14.05\%) & 0.662 (-10.54\%) & 0.578 (-21.89\%) & 0.626 (-15.41\%) & 0.588 (-20.54\%) & 0.620 (-16.22\%) \\ \hline
        KFall            & 0.778             & 0.854 (+9.76\%)       & \textbf{0.902 (+15.94\%)} & 0.870 (+11.84\%) & \textbf{0.902 (+15.94\%)} & 0.794 (+2.06\%) & 0.848 (+8.99\%) & 0.878 (+12.87\%) & 0.822 (+5.65\%) & 0.898 (+15.43\%) & 0.892 (+14.65\%) \\ \hline
        UMAFall          & 0.542             & 0.652 (+20.30\%)      & 0.630 (+16.23\%) & 0.698 (+28.78\%) & 0.712 (+31.18\%)  & 0.575 (+6.09\%) & 0.702 (+29.52\%) & 0.630 (+16.23\%) & \textbf{0.850 (+56.83\%)} & 0.580 (+7.01\%) & 0.756 (+39.48\%) \\ \hline
        SisFall          & 0.732             & 0.748 (+2.19\%)       & 0.742 (+1.37\%) & 0.758 (+3.55\%) & \textbf{0.784 (+7.10\%)}  & 0.752 (+2.73\%) & 0.750 (+2.46\%) & 0.772 (+5.46\%) & 0.776 (+6.01\%) & 0.782 (+6.83\%) & 0.680 (-7.10\%) \\ \hline
    \end{tabular}
    }
\label{tab:performance}
\vspace{-10px}
\end{table}

To determine whether LLM-generated synthetic data improves fall detection performance, we trained and tested the LSTM model with combinations of real and synthetic fall data. Specifically, we utilized the same 28 synthetic datasets generated in the previous section (see Section~\ref{subsect:align_syn_vs_diff}). However, in this analysis, we incorporated the synthetic data into the training sets of the baseline datasets (with 60\% of ADLs, 20\% of real fall samples, and 20\% of synthetic fall samples) and evaluated its impact on model performance. We trained a total of 29 models, including one baseline model trained solely on real data and 28 models trained on a combination of real and synthetic data. We then compared the performance of each model trained with synthetic data against the baseline to assess the effectiveness of data augmentation in improving LSTM performance (see Section~\ref{sect:implement_detaisl} for experimental details). Table~\ref{tab:performance} presents the average F1-scores (for the fall class (class label 1)) of each dataset. 

\textit{For the SMM dataset}, we found that synthetic data did not improve model performance. Every method led to a drop in F1-score, with Diffusion-TS decreasing performance by 8.11\% (0.680 vs. 0.740 baseline). Among text-to-motion models, T2M showed the lowest drop (-4.05\%), while ParCo (-8.11\%) and SATO (-12.43\%) resulted in greater declines. Text-to-text models performed even worse, as GPT4-ZS (-21.89\%) and GPT4o-ZS (-20.54\%) caused the most significant reductions in performance. T2M exhibited the smallest decline among all models, outperforming Diffusion-TS (-8.11\%), which suggests that it generated the most useful synthetic data for this dataset. This is justified by its lower JSD (0.6159) compared to ParCo (0.6265) and SATO (0.7803), indicating better alignment with real data. Among text-to-text models, GPT4-FS (-15.41\%) showed the lowest drop, which aligns with its lower JSD (0.5432) compared to other methods in this category (see Figure~\ref{fig:llms_quantitative_analysis}(a)).

\textit{For KFall dataset}, we observed that synthetic data substantially improved F1-scores across all models. Diffusion-TS improved performance by +9.76\%, while T2M and SATO showed the highest gains (+15.94\%), surpassing ParCo (+11.84\%). Among text-to-text models, GPT4o-ZS (+15.43\%) performed better than GPT4oFS (+14.65\%), indicating a reverse trend where Few-Shot training did not improve alignment and instead resulted in slightly reduced effectiveness. A similar pattern appeared with GPT4, where ZS (+12.87\%) outperformed FS (+5.65\%). T2M achieved the highest gain among text-to-motion models (+15.94\%), which is consistent with its lower JSD (0.5529). Among text-to-text models, GPT4o-ZS (+15.43\%) provided the highest gain, aligning with its lowest JSD (0.4376) among text-to-text models, further confirming its better distributional alignment with real data (see Figure~\ref{fig:llms_quantitative_analysis}(c)).

\textit{For UMAFall dataset}, we found that adding synthetic data led to the most significant improvements in F1-scores. Diffusion-TS increased performance by +20.30\%, while SATO showed the highest gain among text-to-motion models (+31.18\%), followed by ParCo (+28.78\%). Among text-to-text models, GPT4-FS demonstrated the highest overall improvement (+56.83\%), followed by GPT4o-FS (+39.48\%). SATO achieved the highest gain among text-to-motion models, which is justified by its lowest JSD (0.6405) compared to T2M (0.5483) and ParCo (0.6176), suggesting it maintained a better balance between distribution coverage and alignment. Among text-to-text models, GPT4-FS outperformed all other models, including Diffusion-TS (+20.30\%), aligning with its lowest JSD (0.4391) among all models (see Figure~\ref{fig:llms_quantitative_analysis}(e)).

\textit{For SisFall dataset}, we observed moderate improvements from synthetic data, but the gains were smaller compared to KFall and UMAFall. Diffusion-TS improved the F1-score by +2.19\%, while SATO showed the highest gain (+7.10\%) among text-to-motion models, followed by ParCo (+3.55\%). Among text-to-text models, GPT4o-ZS (+6.83\%) achieved the highest improvement, while GPT4-FS (+6.01\%) also provided a notable boost. However, GPT4o-FS was the only model that decreased performance (-7.10\%), confirming the reverse trend we observed in qualitative analysis, where FS settings introduced noise and instability in high-frequency datasets. SATO achieved the highest gain among text-to-motion models, which is consistent with its lowest JSD (0.8128) compared to ParCo (0.7301) and T2M (0.6921). Among text-to-text models, GPT4o-ZS showed the highest gain, aligning with its lowest JSD (0.6021) among all models (see Figure~\ref{fig:llms_quantitative_analysis}(g)), suggesting it preserved a better alignment with real data despite challenges in high-frequency datasets.

\noindent\textbf{Findings:}
Our findings demonstrate that LLM-generated synthetic data can enhance fall detection performance, but its effectiveness varies depending on the characteristics of the baseline dataset. 
For example, for SMM, synthetic data reduced performance, with T2M showing the smallest drop (-4.05\%), however, no method outperformed the baseline. The decline in performance for SMM is likely due to two key factors. First, the baseline dataset characteristics, where data is collected from the left wrist, lead to low-motion variability since the non-dominant wrist captures subtle movements with fewer high fluctuations. Consequently, even few-shot synthetic data trained on the same fall samples reinforced noise rather than enhancing variability. Second, data flow differences between real and synthetic falls contributed to misalignment. The SMM dataset records only the impact phase of falls, whereas synthetic data was generated based on gradual transitions into falls, including steps before losing balance. This discrepancy in fall representation led to poor integration of synthetic data with real samples, reducing model performance.

The largest improvements were observed in UMAFall, where GPT4-FS achieved the highest gain (+56.83\%), surpassing Diffusion-TS (+20.30\%), indicating that FS text-to-text models were particularly effective for this dataset. One reason for performance gain in UMAfall is its limited fall diversity, with only three fall types, making it a less challenging dataset. The addition of synthetic data introduced greater variability, helping the model generalize better and improving the F1-score. Another factor is the low sampling rate (20Hz), which we previously observed to be optimal for text-to-text models, as higher sampling rates led to a loss of precision in synthetic data (see Section~\ref{subsect:align_syn_vs_diff}). This alignment between synthetic and real data contributed to the model’s improved performance.

In KFall, synthetic data enhanced performance, with T2M and SATO achieving the highest gains (+15.94\%), but GPT-4o and GPT4-FS models exhibited a reverse trend, performing worse than ZS. In SisFall, synthetic data provided moderate gains, with SATO (+7.10\%) and GPT-4o Zero-Shot (+6.83\%) performing best. However, GPT-4o Few-Shot showed a reverse trend, decreasing performance (-7.10\%), similar to its behavior in KFall, suggesting instability in high-frequency datasets. One key reason for the performance improvement in KFall and SisFall is that both datasets include fall samples with transitional movements, where the entire sequence, including pre-fall actions like standing, walking, and losing balance, is labeled as a single fall (1) sample. This structure aligned better with the synthetic data, which was generated using prompts that described falls as gradual transitions rather than isolated impact events. However, despite this alignment, synthetic data still exhibited noise and deviations from real data (see Section~\ref{subsect:align_syn_vs_diff}), limiting the overall improvement to marginal gains rather than significant enhancements.

Among generative methods, across all datasets, text-to-motion models (T2M, SATO) consistently outperformed text-to-text models, demonstrating their effectiveness in producing synthetic fall data that better aligns with real data distributions. However, few-shot text-to-text configurations were more effective for lower-frequency datasets (UMAFall) but introduced noise and instability in higher-frequency datasets (SisFall, KFall), limiting their reliability. Diffusion-TS remained a stable and effective choice across all datasets, though it did not always provide the highest gains.

\section{Ablation Study}\label{sect:ablation_study}
To further analyze the impact of LLM-generated synthetic data on fall detection, we conducted an ablation study addressing three key research questions:
\begin{itemize}
    \item Do the baseline dataset characteristics impact the fall detection task using synthetic data?
    \item Does the quantity of LLM-generated data affect the accuracy of fall prediction models?
    \item Which prompting strategy is more effective for generating diverse fall data, human-designed scenarios or self-generated scenarios?
\end{itemize}

\subsection{Do the baseline dataset characteristics impact the fall detection task using synthetic data?}
To validate the impact of baseline dataset characteristics on synthetic data effectiveness, we conducted experiments using UMAFall and SMM, modifying their modalities by extracting waist accelerometer data for UMAFall and right hip accelerometer data for SMM. For text-to-motion models, we extracted synthetic data from the right hip (index 2) for SMM and the waist/pelvic joint (index 0) for UMAFall. From text-to-text models, we used the same zero-shot synthetic data, while to generate few-shot data, we incorporated fall samples from five randomly selected subjects. Additionally, we retrained Diffusion-TS with the updated SMM and UMAFall data to generate synthetic samples with the new modalities. 
After generating synthetic data, we trained and tested the same LSTM model for fall classification using 5-fold cross-validation. The average F1-scores are reported in Table~\ref{tab:mpact_baseline}.
\begin{table}[h!]
    \centering
    \caption{Impact of baseline dataset characteristics on LSTM-based fall detection: Average F1-scores (\(+\)/\(-\)\%) across 5-fold cross-validation.}
    \resizebox{1\textwidth}{!}{ 
    \begin{tabular}{|l|c|c|c|c|c|c|c|c|c|c|c|}
        \hline
        \multirow{3}{*}{\textbf{Dataset}} & \multirow{3}{*}{\textbf{Baseline}} & \multirow{3}{*}{\textbf{Diffusion-TS}} & \multicolumn{3}{c|}{\textbf{Text-to-Motion LLMs}} & \multicolumn{6}{c|}{\textbf{Text-to-Text LLMs}} \\ \cline{4-12} 
                         &                   &                       & \multirow{2}{*}{\textbf{T2M}} & \multirow{2}{*}{\textbf{ParCo}} & \multirow{2}{*}{\textbf{SATO}} & \multicolumn{2}{c|}{\textbf{Gemini-Flash-8B}} & \multicolumn{2}{c|}{\textbf{Copilot}} & \multicolumn{2}{c|}{\textbf{GPT4o}} \\ \cline{7-12} 
                         &                   &                       &              &                &               & \textbf{Zeroshot} & \textbf{Fewshot} & \textbf{Zeroshot} & \textbf{Fewshot} & \textbf{Zeroshot} & \textbf{Fewshot} \\ \hline
        SMM (right hip) & 0.8075 & 0.778 (-3.65\%) & 0.812 (+0.56\%) & \textbf{0.856 (+6.01\%)} & 0.746 (-7.62\%) & 0.762 (-5.63\%) & 0.682 (-15.54\%)  & 0.806 (-0.19\%)  & 0.754 (-6.63\%)  & 0.772 (-4.40\%)   & 0.754 (-6.63\%) \\ \hline
        UMAFall (waist)  & 0.744  & 0.922 (+23.92\%)  & 0.854 (+14.78\%) & 0.850 (+14.25\%) & 0.892 (+19.89\%) & \textbf{0.946 (+27.15\%)}  & 0.710 (-4.57\%)  & 0.888 (+19.35\%) & \textbf{0.946 (+27.15\%)} & 0.888 (+19.35\%)  & 0.702 (-5.65\%) \\ \hline
    \end{tabular}
    }
\label{tab:mpact_baseline}
\end{table}

As shown in Table~\ref{tab:mpact_baseline}, for SMM (right hip), most synthetic datasets failed to improve performance, with several models leading to a decline in F1-scores. Notably, ParCo (+6.01\%) was the only model that achieved a significant improvement, while others such as Diffusion-TS (-3.65\%), SATO (-7.62\%), and text-to-text models exhibited a performance drop. In particular, Gemini-FS (-15.54\%) and GPT4o-FS (-6.63\%) showed the most pronounced declines. Notably, the performance drops in SMM are lower than those observed in Table~\ref{tab:performance}, suggesting that modifying the sensor modality to the right hip improved the alignment between synthetic and real data, mitigating some of the negative impact.

Conversely, in UMAFall (waist), synthetic data significantly improved fall detection performance, with multiple models achieving notable gains. Gemini-ZS and Copilot-FS (+27.15\%) yielded the highest improvements, followed by Diffusion-TS (+23.92\%) and SATO (+19.89\%). This suggests that the synthetic data was better aligned with UMAFall’s real data, likely due to the dataset's simpler motion structure and fewer fall variations. However, GPT-4o Few-Shot (-5.65\%) and Gemini Few-Shot (-4.57\%) resulted in a decline, indicating potential instability in certain text-to-text generated datasets.

\noindent\textbf{Findings:} The results confirm that baseline dataset characteristics significantly influence the effectiveness of synthetic data in fall detection tasks. UMAFall, with its waist-based accelerometer data, lower motion variability, and fewer fall types, benefited from synthetic augmentation, whereas SMM, with right hip data and higher motion complexity, showed limited or negative improvements. These findings suggest that synthetic data is more effective when the baseline dataset has lower complexity and clearer movement patterns but may struggle to align well with datasets that capture more subtle or diverse motion variations.

\subsection{Which prompting strategy is more effective for generating diverse fall data—human-designed scenarios or self-generated scenarios?}
The goal of this study is to assess whether human-designed prompts introduce biases or limitations that confuse LLMs, potentially affecting the diversity and realism of synthetic fall data. By comparing the performance of models trained on self-generated vs. human-designed scenarios, we aim to determine which strategy leads to better variability, improved data alignment in zero-shot settings, and ultimately, stronger fall detection performance.
For this purpose, we conducted an experiment using text-to-text models exclusively. We did not use text-to-motion models, as they are limited to generating individual motion sequences rather than diverse fall scenarios iteratively. To compare human-designed and self-generated fall scenarios, we formulated the following system prompt for LLMs:
\lstset{
  basicstyle=\scriptsize\ttfamily,
  breaklines=true,
  breakatwhitespace=true,
  columns=fullflexible,
  frame=single  
}
\begin{lstlisting}
System Prompt: I want you to create 50 scenarios of persons falling. Each scenario must be unique.
Response: List of 50 textual descriptions of fall scenarios.
System Prompt: Now, for all of these 50 scenarios, create accelerometer data of 4 seconds with x, y, z columns at sampling speed [sampling_rate] Hz. The output CSV must contain 4 columns with names x, y, and z in lower letters. The CSV must be semicolon-separated.
Response: output.CSV
\end{lstlisting}
We then used the synthetic accelerometer data generated from these prompts to train and test an LSTM model over five folds, with the average F1-scores reported in Table~\ref{table:auto_prompts_performance}.
\begin{table}[h!]
\centering
\caption{Comparison of auto-generated and manually designed prompts: Impact on synthetic data quality and fall detection performance.}
\resizebox{0.8\textwidth}{!}{
    \begin{tabular}{|c|c|c|c|c|c|c|c|} \hline
        \multirow{2}{*}{\textbf{Dataset}} & 
        \multirow{2}{*}{\textbf{Baseline}} & 
        \multicolumn{2}{c|}{\textbf{GPT4o}} & 
        \multicolumn{2}{c|}{\textbf{Copilot}} & 
        \multicolumn{2}{c|}{\textbf{Gemini-Flash-8B}} \\ \cline{3-8}
        & & \textbf{Zeroshot} & \textbf{Auto} & \textbf{Zeroshot} & \textbf{Auto} & \textbf{Zeroshot} & \textbf{Auto}  \\ \hline
        SMM (left wrist) & 0.740 & 0.588 (-20.54\%) & \textbf{0.724 (-2.16\%)} & 0.578 (-21.89\%) & 0.714 (-3.51\%) & 0.636 (-14.05\%) & 0.626 (-15.41\%) \\ \hline
        KFall (waist) & 0.778 & \textbf{0.898 (+15.42\%)} & 0.892 (+14.65\%) & 0.878 (+12.85\%) & 0.874 (+12.34\%) & 0.794 (+2.06\%) & 0.850 (+9.25\%)  \\ \hline
        UMAFall (waist) & 0.542 & 0.580 (+7.01\%) & \textbf{0.840 (+54.98\%)} & 0.630 (+16.24\%) & 0.660 (+21.77\%) & 0.575 (+6.09\%)  & 0.723 (+33.39\%) \\ \hline
        SisFall (waist) & 0.732 & \textbf{0.782 (+6.83\%)} & 0.768 (+4.92\%) & 0.772 (+5.46\%) & 0.774 (+5.74\%) & 0.752 (+2.73\%) & 0.772 (+5.46\%) \\ \hline
    \end{tabular}
}
\label{table:auto_prompts_performance}
\vspace{-2px}
\end{table}

For SMM, accelerometer data from the Auto-generated prompts resulted in the least drop in performance across all LLMs compared to human-designed prompts. GPT4o (Auto) achieved 0.724 (-2.16\%), a substantial improvement over GPT4o-ZS (0.588, -20.54\%), which experienced the largest drop in F1-score among all prompting strategies. A similar pattern was observed in GPT4 and Gemini, where Auto led to smaller declines (0.714 and 0.626, respectively) compared to ZS prompting, which suffered larger drops (0.578 and 0.636, respectively). These results suggest that self-generated prompts enable LLMs to construct more diverse and generalizable fall scenarios, mitigating performance declines and leading to better alignment with real-world falls.

For KFall, both ZS and Auto-generated prompts performed well, with minor differences in F1-scores. GPT4o-ZS (0.898, +15.42\%) marginally outperformed Auto-generated prompts (0.892, +14.65\%). A similar trend was observed in GPT4 ZS: 0.878, Auto: 0.874), with Auto performing slightly worse. Gemini, however, showed a larger difference, where Auto-generated prompts improved the F1-score (+9.25\%) compared to ZS (+2.06\%), indicating that Auto-generated prompts were more beneficial for this model. Overall, the prompting strategy had a smaller impact on KFall, likely because the inherent variability of this dataset was already aligned well with synthetic falls, reducing the need for additional prompt diversity.

For UMAFall, Auto-generated prompts resulted in the highest improvements across all models, reinforcing the importance of diverse prompt generation. GPT4o-Auto (0.840, +54.98\%) demonstrated the most substantial increase over GPT4o-ZS (0.580, +7.01\%). Similar trends were observed in GPT4 (Auto: 0.660, +21.77\%) vs. ZS (0.630, +16.24\%), and Gemini (Auto: 0.723, +33.39\%) vs. ZS (0.575, +6.09\%). 
The significant boost in F1-scores for Auto-generated prompts suggests that human-designed scenarios failed to introduce sufficient variability, making the self-generated approach beneficial.

For SisFall, both ZS and Auto-generated prompts led to similar F1-scores, indicating that the prompting strategy had minimal impact on fall detection accuracy. GPT4o-ZS (0.782, +6.83\%) performed slightly better than Auto-generated prompts (0.768, +4.92\%). In contrast, GPT4 (Auto: 0.774, ZS: 0.772) and Gemini (Auto: 0.772, ZS: 0.752) exhibited almost identical performance. These findings suggest that for high-frequency datasets like SisFall, the diversity introduced by Auto-generated prompts was not as beneficial as in lower-frequency datasets like UMAFall.

\noindent\textbf{Findings:}
The results indicate that Auto-generated prompts generally outperform human-designed prompts in generating diverse and realistic fall scenarios, though their impact varies across datasets. For SMM and UMAFall, Auto-generated prompts significantly improved performance, demonstrating that LLMs benefit from self-generated variability rather than manually constrained descriptions. For KFall and SisFall, accelerometer data generated from auto-generated prompts showed minimal improvement. This is likely because these datasets already have intrinsic variability well-aligned with synthetic data, reducing the need for prompt optimization. Similarly, the high sampling rates may have limited the benefits of synthetic data, as LLMs lose specificity regardless of whether prompts are user-designed or auto-generated, highlighting their limitations in high-frequency datasets.

\subsection{Does the quantity of LLM-generated data affect the accuracy of fall prediction models?}
To investigate whether increasing the quantity of LLM-generated synthetic data affects the accuracy of fall prediction models, we conducted an LSTM-based experiment over five folds while systematically adjusting the proportion of synthetic data.  Specifically, in the training sets, we used 50\% ADLs, 10\% real falls to retain natural data, and 40\% synthetic falls, with the same training protocol.

For text-to-motion models, we combined synthetic data from all three models (T2M, ParCo, and SATO) to meet the required percentage, ensuring that a diverse range of generated motions is incorporated into training, reducing the biases that may arise from any single generative approach.
For text-to-text models, we selected the best-performing prompting strategy (ZS or FS) from the previous fall prediction experiments for each baseline dataset. If ZS performed well in the earlier tests for a particular baseline dataset, we used ZS synthetic data, otherwise, we used FS data for that specific dataset. This approach ensured that we leveraged the most effective LLM-generated data for each dataset, rather than simply increasing synthetic data volume without considering quality.
The average F1-scores obtained from these experiments are reported in Table~\ref{table:quantity_of_syn_data}.
\begin{table}[h!]
\centering
\caption{Impact of increasing synthetic data proportion up to 80\% on fall detection performance.}
\resizebox{0.8\textwidth}{!}{
    \begin{tabular}{|l|c|c|c|c|c|} \hline
        \textbf{Dataset} & \textbf{Baseline} & \textbf{Diffusion-TS} & \textbf{T2M+ParCo+SATO} & \textbf{Gemini+GPT4+GPT4o} & \textbf{\makecell{Auto \\ (Gemini+GPT4+GPT4o)}} \\ \hline
        SMM (left wrist)    & 0.740   & 0.680 (-8.11\%)   & 0.730 (-1.35\%) & \textbf{0.746 (+0.81\%)} & 0.626 (-15.41\%) \\ \hline
        KFall (waist)   & 0.778   & 0.854 (+9.77\%)   & 0.846 (+8.74\%) & 0.824 (+5.91\%) & \textbf{0.902 (+15.94\%)} \\ \hline
        UMAFall (right wrist) & 0.542   & 0.652 (+20.30\%)  & 0.680 (+25.46\%) & 0.642 (+18.45\%) & \textbf{0.690 (+27.30\%)} \\ \hline
        SisFall (waist) & 0.732   & 0.748 (+2.19\%)   & 0.744 (+1.64\%) & \textbf{0.762 (+4.10\%)} & 0.738 (+0.82\%)  \\ \hline
    \end{tabular}
}
\label{table:quantity_of_syn_data}
\vspace{-2px}
\end{table}
For SMM, Diffusion-TS (-8.11\%) and Auto (-15.41\%) both degraded performance, indicating that certain synthetic datasets did not generalize well to real fall patterns. Text-to-motion models (T2M+ParCo+SATO) showed a minimal decline (-1.35\%), suggesting that motion-based synthetic data remained relatively stable even at higher proportions. However, text-to-text models (FS Gemini+GPT4+GPT4o) achieved the highest F1-score (+0.81\%), surpassing even the baseline. This improvement aligns with the previous fall prediction tests where FS text-to-text models outperformed ZS models for SMM, indicating that an increased amount of high-quality text-based synthetic data can compensate for limited real data.

For KFall, Auto-synthetic data achieved the highest gain (+15.94\%), outperforming Diffusion-TS (+9.77\%) and T2M+ParCo+SATO (+8.74\%), indicating that the combination of optimized text-to-text models was particularly beneficial. While text-to-text models alone achieved a lower gain (+5.91\%), their contribution in Auto suggests that properly tuned synthetic datasets can significantly enhance real-world fall detection.

For UMAFall, Auto-synthetic data achieved the highest gain (+27.30\%), followed by text-to-motion models (T2M + ParCo + SATO) at +25.46\%. Diffusion-TS (+20.30\%) and text-to-text models (+18.45\%) also showed improvements but to a lesser extent. These results confirm the effectiveness of the earlier FS text-to-text models for UMAFall, where increasing the diversity of synthetic falls improved model generalization.

For SisFall, increasing synthetic data provided moderate improvements, but the gains were smaller compared to KFall and UMAFall. Text-to-text models achieved the highest improvement (+4.10\%), outperforming Diffusion-TS (+2.19\%) and text-to-motion models (+1.64\%). Auto-synthetic data showed only a minimal improvement (+0.82\%), indicating that a higher proportion of synthetic data does not always guarantee better model performance.

\noindent\textbf{Findings:}
The results indicate that increasing synthetic data can improve model performance, but its effectiveness depends on the dataset and the type of synthetic data used. For SMM, text-to-text synthetic data led to a slight improvement, whereas an excessive amount of synthetic motion data did not provide significant benefits. In KFall and UMAFall, increasing synthetic data substantially boosted accuracy, with Auto-generated text-based falls and text-to-motion models yielding the highest gains. However, for SisFall, the performance gains were modest, suggesting that higher proportions of synthetic data may not always benefit models trained on high-frequency datasets. These findings confirm that LLM-generated synthetic data quantity does influence model accuracy, but its impact varies across datasets.


\section{Key Findings and Applicability of LLMs for Fall Detection}\label{sect:key_findings}
After conducting a comprehensive experimental analysis, this section highlights the key findings of our study by addressing the research questions posed in the introduction.
\begin{enumerate}
    \item \textbf{Can LLMs generate accelerometer data specific to gender, age, and joint placement?} \\
    Our results indicate that text-to-motion models (T2M, SATO, ParCo) effectively capture joint-specific variations but fail to differentiate demographic attributes such as gender and age. On the other hand, text-to-text models (GPT4o, GPT4, Gemini) struggled to introduce both joint and demographic variations, producing only minor differences in joint-specific prompts, making them less effective for applications requiring age- or gender-specific biomechanical variations.
    \item \textbf{Do LLM-generated data align more closely with real data than diffusion-based methods?} \\
    When evaluating alignment with real data, Diffusion-TS consistently outperformed all LLM-generated data, achieving the lowest JSD and highest Coverage scores across all datasets. Among LLM-based methods, text-to-motion models (T2M, SATO, ParCo) demonstrated better alignment than text-to-text models, capturing more realistic motion distributions. However, text-to-text models exhibited high variance and spikiness, particularly in datasets with higher sampling rates such as KFall (100Hz) and SisFall (200Hz). This suggests that LLMs lose specificity as the sampling rate increases, leading to poor alignment. Although few-shot prompting slightly improved data realism, it also introduced noise and instability, particularly in high-frequency datasets, further limiting the reliability of LLM-generated synthetic data.
    \item \textbf{Do LLM-generated data improve fall detection model performance?} \\
    The impact of synthetic data on fall detection performance varied depending on the characteristics of the baseline dataset. In SMM, synthetic data caused a decline in performance across all methods, with T2M leading to the smallest reduction (-4.05\%). This decline was primarily due to two factors: first, the left wrist sensor placement, which captures subtle, low-variance movements, resulting in weak fall signal representation; and second, mismatches in data flow, as real data in SMM only includes the impact phase of falls, while synthetic data incorporated transitional movements (e.g., walking, stumbling before impact), leading to poor integration.
    In contrast, UMAFall showed the highest gains, with GPT4-FS improving the F1-score by +56.83\%. This improvement is likely due to the simplicity of the dataset, as it contains only three fall types, making it less challenging for augmentation. Additionally, its low sampling rate (20Hz) aligns well with LLM-generated data, mitigating precision loss that was observed in high-frequency datasets.
    For KFall and SisFall, synthetic data provided moderate performance improvements. However, FS prompting led to instability in high-frequency datasets, with GPT4o-FS reducing F1-score in SisFall (-7.10\%), suggesting that text-to-text models struggle with high-frequency accelerometer data. Text-to-motion models consistently outperformed text-to-text models, while Diffusion-TS remained the most stable method across all datasets, demonstrating its superiority in generating well-aligned synthetic fall data.
\end{enumerate}

\noindent \textbf{Applicability of LLMs for Fall Detection}\\
Despite observed F1-score improvements, the highest gain (56.83\% in UMAFall) remains below 0.9, which is insufficient for real-world deployment. When implemented in wearable devices, quantization further reduces accuracy, making these marginal improvements negligible. Moreover, prompt engineering is labor-intensive and time-consuming, requiring multiple iterations to generate usable data. Even auto-generated prompts lack consistency, and dependency on chatbot APIs introduces variability, making real-time data generation impractical. LLMs offer an alternative for synthetic data generation but lack the reliability and efficiency of diffusion-based methods. Their use in fall detection remains limited, requiring further advancements in automation, consistency, and data refinement for practical application.

A potential alternative to relying on LLMs for synthetic fall data generation is leveraging specialized generative models trained on time-series data. One approach is to develop a time-series foundation model, pre-trained on a vast corpus of human movement data (accelerometer, gyroscope, magnetometer, etc.), capturing a wide range of activities, transitions, and sensor-specific variations. If properly conditioned, such a model could generate fall-specific data tailored to baseline dataset characteristics, including demographic factors like age, gender, and sensor placements. However, training such a model requires massive datasets covering all possible sensor placements and demographic variations, which are difficult to obtain. Moreover, ensuring that the model generates precisely controlled synthetic falls rather than generic motions repurposed from unrelated activities remains a significant challenge.

In contrast, fine-tuning text-to-motion models (like T2M, ParCo, and SATO) provides a more practical and effective solution for fall-specific synthetic data generation. Rather than requiring extensive demographic conditioning, fine-tuning these models on real fall datasets allows them to learn biomechanically accurate fall motions while still maintaining control over sensor placement. Since text-to-motion models inherently generate full-body joint trajectories, we can extract sensor-specific data from the correct joint indices, ensuring alignment with real-world accelerometer signals. This approach avoids the complexity of training a foundation model from scratch and focuses on refining fall-specific motion patterns, making it more feasible, efficient, and targeted for synthetic fall data generation.

Given the challenges in collecting diverse, demographic-specific fall datasets, fine-tuning text-to-motion models emerges as the superior approach. It enables controlled sensor placement, focuses on accurate biomechanical motion, and avoids the impracticality of training a large-scale foundation model from limited fall data. Future research should focus on optimizing these models for fall-specific applications, refining their sensor placement accuracy and biomechanical realism to make synthetic data a truly effective augmentation tool for fall detection systems.

\section{Conclusion and Future Work}\label{concusion_fw}
The motivation behind this study is the challenges in collecting real fall data, especially from elderly individuals, due to ethical and practical constraints. LLMs offer a potential solution by synthesizing fall data that can augment real datasets, potentially enhancing fall detection model performance. However, the effectiveness of synthetic data depends on how well it aligns with real data and its impact on training the LSTM model, which we systematically investigate in this study. 
We generated synthetic accelerometer data using two categories of large language models (LLMs): text-to-motion models (T2M-GPT, SATO, and ParCo) and text-to-text models (GPT4o, GPT4, and Gemini). We then compared the quality and effectiveness of these models against Diffusion-TS-generated data. The text-to-motion models, originally trained on general human motion datasets, were used to extract joint-specific accelerometer data, while the text-to-text models were prompted to directly generate fall-related accelerometer sequences. The impact of synthetic data was evaluated by augmenting real fall datasets and assessing model performance using F1-score as the primary metric.

Our findings reveal that the effectiveness of synthetic data depends on the characteristics of the baseline dataset. While UMAFall showed the highest performance improvement (+56.83\% with GPT4-FS), SMM experienced performance degradation across all synthetic datasets, with T2M causing the smallest drop (-4.05\%). This discrepancy highlights two critical factors: (1) Sensor placement matters, datasets using left wrist sensors (SMM) struggled with synthetic data, while those using waist sensors (UMAFall, KFall) benefited from it, (2) Data flow alignment is crucial, real datasets that included only impact-phase falls (SMM) were poorly complemented by synthetic data that modeled gradual transitions into falls. Additionally, our results demonstrate that text-to-motion models outperformed text-to-text models, as they maintained better alignment with real data, however, Diffusion-TS remained the most stable, but less beneficial method across all datasets for the fall detection task. This is likely because, while its generated samples are well-formed, they lack the necessary diversity and critical fall-specific variations needed to improve model generalization. Additionally, its reliance on learned time-series patterns rather than semantic reasoning limits its ability to simulate the nuanced movement dynamics of real falls, making its synthetic data less impactful for training a robust fall detection model.

An ablation study further confirmed that datasets with low sampling rates (UMAFall, 20Hz) aligned better with LLM-generated data, while high-frequency datasets (SisFall, 200Hz) showed instability with synthetic data due to loss of specificity in LLM-generated data at higher sampling rates. FS prompting helped improve realism but introduced noise, particularly in high-frequency datasets, where GPT4o FS caused a performance drop in SisFall (-7.10\%).

To overcome the limitations of LLMs identified in this study, future research will focus on several key areas. First, we plan to fine-tune text-to-motion models on real fall datasets, ensuring that the generated synthetic data better reflects biomechanical realism (across diverse demographics) and real-world fall characteristics. Additionally, the manual effort required for prompt engineering remains a challenge, making it necessary to explore self-optimizing prompt generation strategies that can automate and refine the process for more efficient synthetic data creation. Another critical area is improving alignment for high-frequency datasets, as models like GPT4o and Gemini exhibited instability when generating synthetic data for SisFall (200Hz). To address this, future work will investigate constraints and optimization techniques that mitigate excessive smoothing or spikiness in high-frequency accelerometer data. Finally, given the dependency on chatbot-based APIs for synthetic data generation, we aim to explore lightweight generative models, specially optimized for edge devices, that can be fine-tuned locally without external API dependencies, ensuring greater control, consistency, and real-time applicability. By addressing these challenges, we aim to enhance the reliability, efficiency, and practicality of synthetic data for fall detection systems, ultimately supporting more robust deployments in wearable devices and healthcare applications.

\section*{Acknowledgments}
This research was supported by the National Science Foundation (NSF) under the Smart and Connected Health (SCH) Program, Grant No. 21223749.

\bibliographystyle{unsrt}  
\bibliography{references}

\end{document}